\documentclass{tlp}

\usepackage{times}
\usepackage{latexsym}
\usepackage{amstext,amssymb}

\newcommand{\ugr}[1]{\ensuremath{\mathit{ugr}(#1)}}

\newcommand{\TBH}[2]{\TPi{#1}{\Pi_{\ref{ex:bh}}}{<}{#2}{\emptyset}}

\newcommand{\WZL}{\textsc{w}}
\newcommand{\DST}{\textsc{d}}
\newcommand{\BE}{\textsc{b}}

\newcommand{\nameset}{{\ensuremath{N}}}

\newtheorem{definition}{Definition}
\newtheorem{theorem}{Theorem}
\newtheorem{corollary}[theorem]{Corollary}
\newtheorem{lemma}{Lemma}[section]

\newcommand{\QED}{\mathproofbox}

 \newenvironment{myproof}[1]{\begin{proof*}[Proof~\ref{#1}]}{\end{proof*}}
 \newenvironment{interproof}[1]{\begin{proof*}[Proof~\ref{#1}]}{\end{proof*}}
 \newenvironment{shortproof}[1]{\begin{xproof}{#1}}{\end{xproof}}

\newcommand{\oksym}{{\mathsf{ok}}}

\newcommand{\blockedsym}{{\mathsf{bl}}}
\newcommand{\appliedsym}{{\mathsf{ap}}}

\newcommand{\ok}[1]{\ensuremath{\oksym(#1)}}
\newcommand{\oko}[2]{\ensuremath{\mathsf{rdy}(#1,#2)}} 

\newcommand{\blocked}[1]{\ensuremath{\blockedsym(#1)}}
\newcommand{\applied}[1]{\ensuremath{\appliedsym(#1)}}

\newcommand{\LPif}{\leftarrow}
\newcommand{\RULE}[3]{#1&=&#2&\LPif&#3}

\newcommand{\nameo}[0]{\mathit{n}}
\newcommand{\namef}[1]{\nameo(#1)}
\newcommand{\name}[1]{\nameo_{#1}}

\newcommand{\head}[1]{\ensuremath{\mathit{head}(#1)}}
\newcommand{\pbody}[1]{\ensuremath{\mathit{body}^{+}(#1)}}
\newcommand{\nbody}[1]{\ensuremath{\mathit{body}^{-}(#1)}}
\newcommand{\body}[1]{\ensuremath{\mathit{body}(#1)}}
\newcommand{\rul}[1]{\ensuremath{\mathit{rule}(#1)}}
\newcommand{\nafo}[0]{\mathit{not}}
\newcommand{\naf}[0]{\nafo\;}

\newcommand{\reduct}[2]{\ensuremath{#1^{#2}}}
\newcommand{\reductr}[1]{\ensuremath{#1^{+}}}
\newcommand{\reductBE}[1]{\ensuremath{#1^{-}}}

\newcommand{\Transo}{\ensuremath{\mathbb{T}}} 
\newcommand{\Trans}[1]{\Transo{(#1)}}
\newcommand{\TransDo}{\Transo^{\DST}}
\newcommand{\TransD}[1]{\TransDo{(#1)}}
\newcommand{\TransWo}{\Transo^{\WZL}}
\newcommand{\TransW}[1]{\TransWo{(#1)}}
\newcommand{\TransBo}{\Transo^{\BE}}
\newcommand{\TransB}[1]{\TransBo{(#1)}}

\newcommand{\To}[1]{\ensuremath{T_{#1}}}
\newcommand{\T}[2]{\To{#1}#2}
\newcommand{\TiO}[2]{\To{#2}^{#1}}
\newcommand{\Ti}[3]{\TiO{#1}{#2}#3}

\newcommand{\TPo}[3]{\ensuremath{\mathcal{T}_{(#1,#2),#3}}}
\newcommand{\TP}[4]{\TPo{#1}{#2}{#3}#4}
\newcommand{\TPX}[4]{\TP{#1}{#2}{#3}{#4}}
\newcommand{\TPXdefault}{\TPX{\Pi}{<}{Y}{X}}
\newcommand{\TPiO}[4]{\TPo{#2}{#3}{#4}^{#1}}
\newcommand{\TPi}[5]{\TPiO{#1}{#2}{#3}{#4}#5}

\newcommand{\TPDo}[3]{\ensuremath{\mathcal{T}^{\DST}_{(#1,#2),#3}}}
\newcommand{\TPD}[4]{\TPDo{#1}{#2}{#3}#4}
\newcommand{\TPDX}[4]{\TPD{#1}{#2}{#3}{#4}}

\newcommand{\TPDiO}[4]{\TPDo{#2}{#3}{#4}^{#1}}
\newcommand{\TPDi}[5]{\TPDiO{#1}{#2}{#3}{#4}#5}

\newcommand{\TPWo}[3]{\ensuremath{\mathcal{T}^{\WZL}_{(#1,#2),#3}}}
\newcommand{\TPW}[4]{\TPWo{#1}{#2}{#3}#4}
\newcommand{\TPWX}[4]{\TPW{#1}{#2}{#3}{#4}}

\newcommand{\TPWiO}[4]{\TPWo{#2}{#3}{#4}^{#1}}
\newcommand{\TPWi}[5]{\TPWiO{#1}{#2}{#3}{#4}#5}

\newcommand{\TPBo}[3]{\ensuremath{\mathcal{T}^{\BE}_{(#1,#2),#3}}}
\newcommand{\TPB}[4]{\TPBo{#1}{#2}{#3}#4}
\newcommand{\TPBX}[4]{\TPB{#1}{#2}{#3}{#4}}
\newcommand{\TPBXdefault}{\TPBX{\Pi}{<}{Y}{X}}
\newcommand{\TPBiO}[4]{\TPBo{#2}{#3}{#4}^{#1}}
\newcommand{\TPBi}[5]{\TPBiO{#1}{#2}{#3}{#4}#5}

\newcommand{\CnO}[0]{\ensuremath{\mathit{Cn}}}
\newcommand{\Cn}[1]{\ensuremath{\CnO(#1)}}
\newcommand{\CPO}[2]{\ensuremath{\mathcal{C}_{(#1,#2)}}}
\newcommand{\CPX}[3]{\ensuremath{\CPO{#1}{#2}(#3)}}
\newcommand{\CPOdefault}{\CPO{\Pi}{<}}
\newcommand{\CPXdefault}{\CPX{\Pi}{<}{X}}

\newcommand{\CPDO}[2]{\ensuremath{\mathcal{C}^{\DST}_{(#1,#2)}}}
\newcommand{\CPDX}[3]{\ensuremath{\CPDO{#1}{#2}(#3)}}
\newcommand{\CPDOdefault}{\CPDO{\Pi}{<}}
\newcommand{\CPDXdefault}{\CPDX{\Pi}{<}{X}}

\newcommand{\CPWO}[2]{\ensuremath{\mathcal{C}^{\WZL}_{(#1,#2)}}}
\newcommand{\CPWX}[3]{\ensuremath{\CPWO{#1}{#2}(#3)}}
\newcommand{\CPWOdefault}{\CPWO{\Pi}{<}}
\newcommand{\CPWXdefault}{\CPWX{\Pi}{<}{X}}

\newcommand{\CPBO}[2]{\ensuremath{\mathcal{C}^{\BE}_{(#1,#2)}}}
\newcommand{\CPBX}[3]{\ensuremath{\CPBO{#1}{#2}(#3)}}
\newcommand{\CPBOdefault}{\CPBO{\Pi}{<}}
\newcommand{\CPBXdefault}{\CPBX{\Pi}{<}{X}}

\newcommand{\PREC}[2]{#1\prec #2}

\newcommand{\PRECMo}[0]{<}
\newcommand{\PRECM}[2]{\ensuremath{#1\PRECMo#2}}

\newcommand{\GR}[2]{\ensuremath{{\Gamma}_{#1}{#2}}}

\newcommand{\ap}[2]   {\ensuremath{{a_{#1}(#2   )}}} 
\newcommand{\bl}[3]   {\ensuremath{{b_{#1}(#2,#3)}}} 
\newcommand{\cokt}[2] {\ensuremath{{c_{#1}}(#2)}}    
\newcommand{\cok}[3]  {\ensuremath{{c_{#1}(#2,#3)}}} 

\newcommand{\Lit}[0]{\ensuremath{\mathit{Lit}}}
\newcommand{\Atm}[0]{\ensuremath{\mathit{At}}}

\title{A semantic framework for preference handling in answer set programming}

\author[Torsten Schaub and Kewen Wang]{%
  Torsten Schaub\thanks{\ Affiliated with the
                          School of Computing Science at
                          Simon Fraser University,
                          Burnaby, Canada.}
                        \\
  Institut f\"ur Informatik,
  Universit\"at Potsdam      \\  
  Postfach 60 15 53,
  D--14415 Potsdam,
  Germany                    \\
  torsten@cs.uni-potsdam.de
  \and
  Kewen Wang\thanks{\ This work was done while the
                      second author was with the 
                      University of Potsdam.}
                    \\
  School of Computing and Information Technology\\
  Griffith University,
  Brisbane 4111,
  Australia\\
  K.Wang@cit.gu.edu.au
}

\begin{document}

\maketitle
\begin{abstract}
We provide a semantic framework for preference handling in answer set
programming. 
To this end,
we introduce preference preserving consequence operators. 
The resulting fixpoint characterizations provide us with a uniform semantic
framework for characterizing preference handling in existing approaches. 
Although our approach is extensible to other semantics by means of an
alternating fixpoint theory, we focus here on the elaboration of preferences
under answer set semantics.
Alternatively, we show how these approaches can be characterized by the concept
of order preservation.
These uniform semantic characterizations provide us with new insights about
interrelationships and moreover about ways of implementation.
\end{abstract}



\section{Introduction}
\label{sec:introduction}

Preferences constitute a very natural and effective way of resolving
indeterminate situations.
For example, in scheduling not all deadlines may be simultaneously
satisfiable,
and in configuration various goals may not be simultaneously met.
In legal reasoning, laws may apply in different situations, but laws may also
conflict with each other.
In fact,
while logical preference handling constitutes already an indispensable means
for legal reasoning systems (cf.~\cite{gordon93a,prakken97a}),
it is also advancing in other application areas
such as
intelligent agents and e-commerce \cite{grosof99a}
and
the resolution of grammatical ambiguities~\cite{cuiswi01}.
The growing interest in preferences is also reflected by the
large number of proposals in logic programming
\cite{sakino96,brewka96a,gelson97a,zhafoo97a,grosof97,breeit99a,descto00c,wazhli00}.
A common approach is to employ meta-formalisms for characterizing
``preferred answer sets''.
This has led to a diversity of approaches that are hardly comparable
due to considerably different ways of formal characterization.
Hence, there is no homogeneous account of preference.

We address this shortcoming by proposing a uniform semantical framework for
extended logic programming with preferences.
To be precise, we develop an (alternating) fixpoint theory for so-called 
\emph{ordered logic programs} (also, prioritized logic programs).
An {ordered} logic program is an extended logic program whose rules are
subject to a strict partial order.
In analogy to standard logic programming, such a program is then interpreted
by means of an associated fixpoint operator.
We start by elaborating upon a specific approach to preference handling 
that avoids some problems of related approaches.
We also show how the approaches of Brewka and Eiter~\shortcite{breeit98a} and
Delgrande et al.~\shortcite{descto00c} can be captured within our framework.
As a result, we obtain that the investigated approaches yield an
increasing number of answer sets depending on how ``tight'' they integrate
preferences.
For obtaining a complementary perspective,
we also provide characterizations in terms of the property
of order preservation, originally defined in~\cite{descto00c} for
distinguishing ``preferred'' from ``non-preferred'' answer sets.
Moreover, we show how these approaches can be implemented by the
compilation techniques developed in~\cite{descto00c}.
As well, we show that all these different preferred answer set semantics
correspond to the perfect model semantics on stratified programs.
We deal with approaches whose preferred answer sets semantics amounts
to a selection function on the standard answer sets of an ordered logic program.
In view of our interest in compiling these approaches into ordinary logic
programs, we moreover limit our investigation to those guaranteeing
polynomial translations.
This excludes approach like the ones in \cite{rintanen95,zhafoo97a} that step
outside the complexity class of the underlying logic programming framework.
This applies also to the approach in~\cite{sakino96},
where preferences on literals are investigated.
While the approach of~\cite{gelson97a} remains within NP, it advocates
strategies that are non-selective 
(as discussed in Section~\ref{sec:discussion}).
Approaches that can be addressed within this framework include those in
\cite{baahol93a,brewka94a} that were originally proposed for default logic.

{The paper is organized as follows.}
Once Section~\ref{sec:background} has provided formal preliminaries,
we begin in
Section~\ref{sec:fp} by elaborating upon our initial semantics for ordered
logic programs.
Afterwards, we show in Section~\ref{sec:fp:others} how this semantics has to
be modified in order to account for the two other aforementioned approaches.



\section{Definitions and notation}
\label{sec:background}

We assume a basic familiarity with alternative semantics of logic
programming~\cite{lifschitz96a}.
An \emph{extended logic program} is a finite set of rules of the form
\begin{equation}\label{eqn:rule}
L_0\LPif L_1,\dots,L_m,\naf L_{m+1},\dots,\naf L_n,
\end{equation}
where $n\geq m\geq 0$, and each $L_i$ $(0\leq i\leq n)$ is a \emph{literal},
ie.\ either an atom $A$ or the negation $\neg A$ of $A$.
The set of all literals is denoted by $\Lit$.
Given a rule $r$ as in~(\ref{eqn:rule}),
we let $\head{r}$ denote the \emph{head}, $L_0$, of $r$
and
$\body{r}$ the \emph{body},
\(
\{L_1,\dots,L_m,\ \naf L_{m+1},\dots,\naf L_n\}
\),
of $r$.
Further, let
\(
\pbody{r}
=
\{L_1,,\dots, L_m\}
\)
and
\(
\nbody{r}
=
\{L_{m+1},\dots, L_n\}
\).
A program is called \emph{basic} if $\nbody{r}=\emptyset$ for all its rules;
it is called \emph{normal} if it contains no classical negation symbol $\neg$.
The reduct of a rule $r$ is defined as $\reductr{r}=\head{r}\LPif\pbody{r}$;
the \emph{reduct}, $\reduct{\Pi}{X}$, of a program $\Pi$ \emph{relative to} a
set $X$ of literals is defined by
\[
\reduct{\Pi}{X}
=
\{\reductr{r}\mid r\in\Pi\text{ and }\nbody{r}\cap X=\emptyset\}.
\]
A set of literals $X$ is \emph{closed under} a basic program $\Pi$ iff for any
$r\in\Pi$, $\head{r}\in X$ whenever $\pbody{r}\subseteq X$.
We say that $X$ is \emph{logically closed} iff it is either consistent 
(ie.\ it does not contain both a literal $A$ and its negation $\neg A$) or
equals $\Lit$.
The smallest set of literals which is both logically closed and closed under a
basic program $\Pi$ is denoted by $\Cn{\Pi}$.
With these formalities at hand,
we can define \emph{answer set semantics} for extended logic programs:
A set $X$ of literals is an \emph{answer set} of a program $\Pi$
iff
$\Cn{\reduct{\Pi}{X}}=X$.
For the rest of this paper, we concentrate on \emph{consistent} answer sets.
For capturing other semantics, $\Cn{\Pi^X}$ is sometimes regarded as
an operator $C_{\Pi}(X)$.
The anti-monotonicity of $C_{\Pi}$ implies that $C^2_{\Pi}$ is monotonic. 
As shown in~\cite{vangelder93}, 
different semantics are obtained by distinguishing different groups of
(alternating) fixpoints of $C^2_{\Pi}(X)$.

Alternative inductive characterizations for
the operators \CnO\ and $C_\Pi$ can be obtained by appeal to
\emph{immediate consequence operators}~\cite{lloyd87}.
Let $\Pi$ be a basic program and $X$ a set of literals.
The \emph{immediate consequence operator} $\To{\Pi}$ is defined as follows:
\[
\T{\Pi}{X} = \{\head{r}\mid r\in\Pi\text{ and }\body{r}\subseteq X\}
\]
if $X$ is consistent, and $\T{\Pi}{X} = \Lit$ otherwise.
Iterated applications of $\To{\Pi}$ are written as $\TiO{j}{\Pi}$ for
$j\geq 0$, where
\(
\Ti{0}{\Pi}{X}=X
\)
and
\(
\Ti{i}{\Pi}{X}=\T{\Pi}{\Ti{i-1}{\Pi}{X}}
\)
for $i\geq 1$.
It is well-known that
\(
\Cn{\Pi}=\bigcup_{i\geq 0}\Ti{i}{\Pi}{\emptyset}
\),
for any basic program $\Pi$.
Also, for any answer set $X$ of program $\Pi$, it holds that
\(
X=\bigcup_{i\geq 0}\Ti{i}{\reduct{\Pi}{X}}{\emptyset}
\).
A reduction from extended to basic programs is avoidable with an extended
operator:
Let $\Pi$ be an extended program and $X$ and $Y$ be sets of literals.
The \emph{extended immediate consequence operator} \To{\Pi,Y} is defined as
follows:
\begin{equation}
  \label{eq:def:extended:immediate:consequence:operator}
  \T{\Pi,Y}{X}
  =
  \{\head{r}\mid r\in\Pi
                 , 
                 \pbody{r}\subseteq X,
                 \text{ and }
                 \nbody{r}\cap Y=\emptyset
  \}
\end{equation}
if $X$ is consistent, and $\T{\Pi,Y}{X} = \Lit$ otherwise.
Iterated applications of $\To{\Pi,Y}$ are written as those of $\To{\Pi}$.
Clearly, we have
\(
\T{\Pi,\emptyset}{X}=\T{\Pi}{X}
\)
for any basic program $\Pi$
and
\(
\T{\Pi,Y}{X}=\T{\Pi^Y}{X}
\)
for any extended program $\Pi$.
Accordingly, we have for any answer set $X$ of program $\Pi$ that
\(
X=\bigcup_{i\geq 0}\Ti{i}{\Pi,X}{\emptyset}
\).
Finally,
for dealing with the individual rules
in~(\ref{eq:def:extended:immediate:consequence:operator}),
we rely on the notion of \emph{activeness}:
Let $X,Y\subseteq\Lit$ be two sets of literals in a program $\Pi$.
A rule $r$ in $\Pi$ is \emph{active} wrt the pair $(X,Y)$,
if $\pbody{r}\subseteq X$ and $\nbody{r}\cap Y=\emptyset$.
Alternatively, we thus have that
\(
\T{\Pi,Y}{X}
=
\{\head{r}\mid r\in\Pi\text{ is active wrt }(X,Y)\}
\).

Lastly,
an \emph{ordered logic program} is simply a pair $(\Pi,<)$, where $\Pi$ is an
extended logic program and $<\;\subseteq{\Pi\times\Pi}$ is an irreflexive and
transitive relation.
Given, $r_1,r_2\in\Pi$, the relation $r_1<r_2$ is meant to express that $r_2$
has \emph{higher priority} than $r_1$.
Programs associated with such an external ordering are also referred to as
\emph{statically} ordered programs,
as opposed to \emph{dynamically} ordered programs whose order relation is
expressed through a special-purpose predicate within the program.



\section{Preferred fixpoints}
\label{sec:fp}

We elaborate upon a semantics for ordered logic program that allows us to
distinguish the ``preferred'' answer sets of a program $(\Pi,<)$ by means of
fixpoint equations.
That is, a set of literals $X$ is a preferred answer set of $(\Pi,<)$,
if it satisfies the equation
\(
\CPXdefault=X
\)
for some operator \CPOdefault.
In view of the classical approach described above,
this makes us investigate semantics that interpret preferences as inducing
selection functions on the set of standard answer sets of the underlying
non-ordered program $\Pi$.

Answer sets are defined via a reduction of extended logic programs to basic
programs.
Controlling such a reduction by means of preferences is difficult
since all conflicts are simultaneously resolved when turning $\Pi$ into
$\reduct{\Pi}{X}$.
Furthermore,
we argue that conflict resolution must be addressed among the original rules in
order to account for blockage between rules.
In fact, once the negative body \nbody{r} is eliminated there is no way to
detect whether $\head{r'}\in\nbody{r}$ holds in case of $r<r'$.
Our idea is thus to characterize preferred answer sets by an inductive
development that agrees with the given ordering.
In terms of a standard answer set $X$, this means that we favor its formal
characterization as
\(
X=\bigcup_{i\geq 0}\Ti{i}{\Pi,X}{\emptyset}
\) 
over $X=\Cn{\reduct{\Pi}{X}}$.
This leads us to the following definition.
%
\begin{definition}\label{def:Tp:W}
Let $(\Pi,<)$ be an ordered logic program and let $X$ and $Y$ be sets of literals.

We define the set of 
immediate consequences of $X$ with respect to $(\Pi,<)$ and $Y$
as
\[
\TP{\Pi}{<}{Y}{X}
\quad=\quad
\left\{
  \head{r}\left|\;
    \begin{array}{rl}
      \mathit{I}. & r\in\Pi
                    \text{ is active wrt } (X,Y)
                    \text{ and}
      \\
      \mathit{II}. & \text{there is no rule }r'\in\Pi
                     \text{ with } r< r'
                     \\ 
                   & \text{such that}
                     \\
                   & (a)\ r'
                        \text{ is active wrt } (Y,X)
                        \text{ and}
                        \\
                   & (b)\ \head{r'}\not\in X
    \end{array}
  \right\}\right.
\]

if $X$ is consistent, and $\TPX{\Pi}{<}{Y}{X}=\Lit$ otherwise.
\end{definition}
%
Note that \TPo{\Pi}{<}{Y} is a refinement of its classical counterpart
\To{\Pi,Y} in~(\ref{eq:def:extended:immediate:consequence:operator}).
The idea behind Condition~\emph{II} is to apply a rule $r$ only if the
``question of applicability'' has been settled for all higher-ranked rules
$r'$.
Let us illustrate this in terms of iterated applications of $\TPo{\Pi}{<}{Y}$.
In this case, $X$ accumulates conclusions, 
while $Y$ comprises the putative answer set.
Then, the ``question of applicability'' is considered to be settled for a higher
ranked rule $r'$
\begin{itemize}
\item if the prerequisites of $r'$ will never be derivable,
viz.\ $\pbody{r'}\not\subseteq Y$,
\quad or
\item if $r'$ is defeated by what has been derived so far,
viz.\ $\nbody{r}\cap X\neq\emptyset$,
\quad or 
\item if $r'$ or another rule with the same head have already applied,
viz.\ $\head{r'}\in X$.
\end{itemize}
The first two conditions show why activeness of $r'$ is stipulated wrt
$(Y,X)$, as opposed to $(X,Y)$ in Condition~\textit{I}.
The last condition serves two purposes:
First, it detects whether the higher ranked rule $r'$ has applied and,
second, it suspends the preference $r<r'$ whenever the head of the higher
ranked has already been derived by another rule.
This suspension of preference constitutes a distinguishing feature of the
approach at hand.

As with $\To{\Pi,Y}$,
iterated applications of $\TPo{\Pi}{<}{Y}$ are written as
$\TPiO{j}{\Pi}{<}{Y}$ for $j\geq 0$, where
\(
\TPi{0}{\Pi}{<}{Y}{X}=X
\)
and
\(
\TPi{i}{\Pi}{<}{Y}{X}=\TP{\Pi}{<}{Y}{\TPi{i-1}{\Pi}{<}{Y}{X}}
\)
for $i\geq 1$.
The counterpart of operator $C_\Pi$ for ordered programs is then defined as
follows.
%
\begin{definition}\label{def:fixpoint:operator:W}
Let $(\Pi,<)$ be an ordered logic program and let $X$ be a set of literals.

We define
\(
\CPXdefault=\bigcup_{i\geq 0}\TPi{i}{\Pi}{<}{X}{\emptyset}
\).
\end{definition}
%
Clearly, \CPO{\Pi}{<} is a refinement of $C_\Pi$.
The difference is that \CPO{\Pi}{<} obtains consequences directly from $\Pi$
and $Y$,
while $C_\Pi$ (normally) draws them by appeal to  \CnO\ after reducing $\Pi$ to
\reduct{\Pi}{Y}.
All this allows us to define preferred answer sets as fixpoints of \CPOdefault.
%
\begin{definition}\label{def:fixpoint:W}
Let $(\Pi,<)$ be an ordered logic program and let $X$ be a set of literals.

We define $X$ as a preferred answer set of $(\Pi,<)$ iff
\(
\CPXdefault=X
\).
\end{definition}

For illustration,
consider the following ordered logic program $(\Pi_{\ref{ex:bh}},<)$:
\begin{equation}
  \label{ex:bh}
  \begin{array}[t]{crcl}
    r_1: & \neg f & \leftarrow & p, \naf f \\
    r_2: & w & \leftarrow & b, \naf \neg w \\
    r_3: & f & \leftarrow & w, \naf \neg f
  \end{array}
  \qquad
  \begin{array}[t]{crcl}
    r_4: & b & \leftarrow & p \\
    r_5: & p & \leftarrow &
  \end{array}
  \qquad\qquad
  r_2< r_1
\end{equation}
Observe that $\Pi_{\ref{ex:bh}}$ admits two answer sets: 
$X=\{p, b, \neg f, w\}$ and $X'=\{p, b, f, w\}$.
As argued in~\cite{baahol93a}, $X$ is preferred to $X'$.
To see this, observe that
\begin{equation}\label{ex:bh:Tp}
\begin{array}[c]{rclp{5mm}rcl}
\TBH{0}{X}&=&\emptyset               &&   \TBH{0}{X'}&=&\emptyset \\
\TBH{1}{X}&=&\{p\}                   &&   \TBH{1}{X'}&=&\{p\} \\
\TBH{2}{X}&=&\{p,b,\neg f\}          &&   \TBH{2}{X'}&=&\{p,b\}\\
\TBH{3}{X}&=&\{p,b,\neg f,w\}        &&   \TBH{3}{X'}&=&\TBH{2}{X'}\\
\TBH{4}{X}&=&\TBH{3}{X}=X            &&            &\neq&X'
\end{array}
\end{equation}
We thus get
\(
\CPX{\Pi_{\ref{ex:bh}}}{<}{X}
=
X
\),
while
\(
\CPX{\Pi_{\ref{ex:bh}}}{<}{X'}
=
\{p,b\}\neq X'
\).
Note that $w$ cannot be included into $\TBH{3}{X'}$ since $r_1$ is active wrt
$(X',\TBH{2}{X'})$ and $r_1$ is preferred to $r_2$.

It is important to see that preferences may sometimes be too strong and deny
the existence of preferred answer sets although standard ones exist.
This is because preferences impose additional dependencies among rules that
must be respected by the resulting answer sets.
This is nicely illustrated by programs $\Pi_{\ref{ex:incoherence}}=\{r_1,r_2\}$ and $\Pi'_{\ref{ex:incoherence}}=\{r'_1,r'_2\}$,
respectively:
\begin{equation}\label{ex:incoherence}
  \begin{array}[c]{rcrcl}
    r_1 & = & a & \leftarrow &      b \\
    r_2 & = & b & \leftarrow &        
  \end{array}
  \qquad\qquad\qquad
  \begin{array}[c]{rcrcl}
    r'_1 & = & a & \leftarrow & \naf b \\
    r'_2 & = & b & \leftarrow &        
  \end{array}
\end{equation}
Observe that in $\Pi_{\ref{ex:incoherence}}$ rule $r_1$ depends $r_2$,
while in $\Pi'_{\ref{ex:incoherence}}$ rule $r'_1$ is defeated by $r'_2$.
But despite the fact that $\Pi_{\ref{ex:incoherence}}$ has answer set $X=\{a,b\}$ and
$\Pi'_{\ref{ex:incoherence}}$ has answer set $X'=\{b\}$,
we obtain no preferred answer set after imposing preferences $r_2< r_1$ and
$r'_2<'r'_1$, respectively.
To see this, observe that
\newcommand{\TINC}[2]{\TPi{#1}{\Pi_{\ref{ex:incoherence}}}{<}{#2}{\emptyset}}
\newcommand{\TINCp}[2]{\TPi{#1}{\Pi'_{\ref{ex:incoherence}}}{<'}{#2}{\emptyset}}
\(
\TINC {0}{X }=\TINC {1}{X }=\emptyset\neq X 
\)
and
\(
\TINCp{0}{X'}=\TINCp{1}{X'}=\emptyset\neq X'
\).
In both cases, the preferred rules $r_1$ and $r'_1$, respectively, are (initially)
inapplicable:
$a\leftarrow     b$ is not active wrt $(\emptyset,\{a,b\})$
and
$a\leftarrow\naf b$ is not active wrt $(\emptyset,\{  b\})$.
And the application of the second rule $b\leftarrow$ is inhibited by Condition~II:
In the case of~$\TINC{1}{X}$, rule $a\leftarrow b$ is active wrt
$(\{a,b\},\emptyset)$; informally, $X$ puts the construction on the false
front that $b$ will eventually be derivable.
In the case of~$\TINC{1}{X}$, rule $a\leftarrow\naf b$ is active wrt
$(\{b\},\emptyset)$.
This is due to the conception that a higher-ranked rule can never be defeated
by a lower-ranked one.

\paragraph{Formal elaboration.}
\label{sec:fixpoint:properties}
We start with the basic properties of our consequence operator:
%
\begin{theorem}\label{thm:TP:properties}
  Let $(\Pi,<)$ be an ordered program
  and
  let $X$ and $Y$ be sets of literals.
  Then, we have:
  \begin{enumerate}
  \item $\TPXdefault\subseteq \T{\Pi,Y}{X}$.
  \item $\TPX{\Pi}{\emptyset}{Y}{X}=\T{\Pi,Y}{X}$.
  \end{enumerate}
  For $i=1,2$, 
  let $X_i$ and $Y_i$ be sets of literals
  and
  ${<_i}\subseteq{\Pi\times\Pi}$ be strict partial orders.
  \begin{enumerate}\addtocounter{enumi}{2}
  \item If $X_1\subseteq X_2$,
    then $\TPX{\Pi}{<}{Y}{X_1}\subseteq\TPX{\Pi}{<}{Y}{X_2}$.
  \item If $\;Y_1\subseteq Y_2,\;$
    then $\TPX{\Pi}{<}{Y_2}{X}\subseteq\TPX{\Pi}{<}{Y_1}{X}$.
  \item If ${<_1}\subseteq{<_2}$,
    then $\TPX{\Pi}{<_2}{Y}{X}\subseteq\TPX{\Pi}{<_1}{Y}{X}$.
  \end{enumerate}
\end{theorem}
%
The next results show how our fixpoint operator relates to its classical
counterpart.
%
\begin{theorem}\label{thm:CP:properties}
  Let $(\Pi,<)$ be an ordered program
  and
  let $X$ be a set of literals.
  Then, we have:
  \begin{enumerate}
  \item $\CPXdefault\subseteq C_\Pi(X)$.
  \item $\CPXdefault=         C_\Pi(X)$, if $X\subseteq\CPXdefault$.
  \item  $\CPX{\Pi}{\emptyset}{X}=C_\Pi(X)$.
\end{enumerate}
\end{theorem}
%
We obtain the following two corollaries.
%
\begin{corollary}\label{thm:CP:Cp}
  Let $(\Pi,<)$ be an ordered logic program
  and
  $X$ a set of literals.

  If   $X$ is a  preferred answer set of\/ $(\Pi,<)$, 
  then $X$ is an           answer set of\/  $\Pi$.
\end{corollary}
%
Our strategy thus implements a selection function among the standard answer
sets of the underlying program.
This selection is neutral in the absence of preferences, as shown next.
%
\begin{corollary}\label{thm:CP:empty}
  Let $\Pi$ be a logic program
  and
  $X$ a set of literals.

  Then,
  $X$ is a  preferred answer set of\/ $(\Pi,\emptyset)$
  iff
  $X$ is an           answer set of\/  $\Pi$.
\end{corollary}
%
Of interest in view of an alternating fixpoint theory is that
\CPOdefault\ enjoys \emph{anti-monotonicity}:
%
\begin{theorem}\label{thm:CP:anti:monotonicty}
  Let $(\Pi,<)$ be an ordered logic program
  and
  $X_1,X_2$ sets of literals.

  If $X_1\subseteq X_2$,
  then
  \(
  \CPX{\Pi}{<}{X_2}\subseteq\CPX{\Pi}{<}{X_1}
  \).  
\end{theorem}

We next show that for any answer set $X$ of a program $\Pi$, there is an
ordering $<$ on the rules of $\Pi$ such that $X$ is the unique preferred
answer set of $(\Pi,<)$.
%
\begin{theorem}\label{thm:total:uniqueness:W}
  Let $\Pi$ be a logic program and $X$ an answer set of $\Pi$. 
  Then,
  there is a strict partial order $<$ such that $X$ is the 
  unique preferred answer set of 
  the ordered program $(\Pi,<)$.
\end{theorem}
%
Our last result shows that a total order selects at most one standard answer
set.
%
\begin{theorem}\label{thm:total:at:most:one:W}
  Let $(\Pi,\ll)$ be an ordered logic program
  and
  $\ll$ be a total order.

  Then, $(\Pi,\ll)$ has zero or one preferred answer set.
\end{theorem}

\paragraph{Relationship to perfect model semantics.}

Any sensible semantics for logic programming should yield, in one
fashion or other, the smallest Herbrand model \Cn{\Pi} whenever $\Pi$ is
a basic program.
A similar consensus seems to exist regarding the \emph{perfect model semantics}
of \emph{stratified} normal programs~\cite{apblwa87,przymusinski88a}.
Interestingly, stratified programs can be associated with a rule ordering in a
canonical way.
We now show that our semantics corresponds to the perfect model semantics on
stratified normal programs.

A normal logic program $\Pi$ is \emph{stratified},
if $\Pi$ has a partition, called \emph{stratification},
\(
\Pi=\Pi_1\cup\dots\cup\Pi_n
\)
such that the following conditions are satisfied for $i,j\in\{1,\dots,n\}$:
\begin{enumerate}
\item $\Pi_i\cap \Pi_j=\emptyset$ for $i\neq j$\/;
\item
  \(
  \pbody{r}\cap(\bigcup_{k=i+1}^n\head{\Pi_k})=\emptyset
  \)
  and
  \(
  \nbody{r}\cap\left(\bigcup_{k=i  }^n\head{\Pi_k}\right)=\emptyset
  \) 
  for all $r\in\Pi_i$.
\end{enumerate}
That is,
whenever a rule $r$ belongs to $\Pi_i$,
      the atoms in $\pbody{r}$ can only appear in the heads of $\bigcup_{k=1}^i    \Pi_k$,
while the atoms in $\nbody{r}$ can only appear in the heads of $\bigcup_{k=1}^{i-1}\Pi_k$.

A stratification somehow reflects an intrinsic order among the rules of a
program.
In a certain sense, rules in lower levels are preferred over rules in higher
levels,
insofar as rules in lower levels should be considered before rules in higher
levels.
Accordingly, the intuition behind the perfect model of a stratified program is
to gradually derive atoms, starting from the most preferred rules.
Specifically, one first applies the rules in $\Pi_1$, resulting in a set of
atoms $X_1$;
then one applies the rules in $\Pi_2$ relative to the atoms in $X_1$;
and so on.

Formally,
the \emph{perfect model semantics} of a stratified logic program
\(
\Pi=\Pi_1\cup\dots\cup\Pi_n
\)
is recursively defined for $0< i< n$ as
follows~\cite{apblwa87,przymusinski88a}.
\begin{enumerate}
\item $X_0=\emptyset$
\item
  \(
  X_{i+1}
  =
  \bigcup_{j\geq 0}\Ti{j}{\Pi_{i+1},X_i}{X_i}
  \)
\end{enumerate}
The \emph{perfect model} $X^\star$ of $\Pi$ is then defined as
\(
X^\star=X_n
\).

Let $\Pi$ be a stratified logic program and $\Pi=\Pi_1\cup\dots\cup\Pi_n$ be a
stratification of $\Pi$.
A natural priority relation $<_s$ on $\Pi$ can be defined as follows:
\[ 
  \text{ For any }
  r_1,r_2\in\Pi,
  \text{ we define }
  r_1<_s r_2
  \text{ iff }
  r_1\in \Pi_i \text{ and } r_2\in \Pi_j
  \text{ such that }
  j<i
  \ .   
\] 
That is, $r_2$ is preferred to $r_1$ if the level of $r_2$ is lower than that
of $r_1$.
We obtain thus an ordered logic program $(\Pi,<_s)$ for any stratified logic
program $\Pi$ with a fixed stratification.
%
\begin{theorem}\label{thm:stratified:perfect:W:i}
  Let $X^\star$ be the perfect model of stratified logic program $\Pi$
  and 
  let $<_s$ be an order induced by some stratification of $\Pi$.
  Then, we have 
  \begin{enumerate}
  \item $X^\star=\CPX{\Pi}{<_s}{X^\star}$,
  \item If $X\subseteq\CPX{\Pi}{<_s}{X}$, then $X^\star=X$.
  \end{enumerate}
\end{theorem}
%
These results imply the following theorem.
%
\begin{corollary}\label{thm:stratified:perfect:W:ii}
  Let $X^\star$ be the perfect model of stratified logic program $\Pi$
  and 
  let $<_s$ be an order induced by some stratification of $\Pi$.
  Then , $(\Pi,<_s)$ has the unique preferred answer set $X^\star$.
\end{corollary}

Interestingly, both programs $\Pi_{\ref{ex:incoherence}}$ as well as
$\Pi'_{\ref{ex:incoherence}}$ are stratifiable.
None of the induced orderings, however, contains the respective preference
ordering imposed in~(\ref{ex:incoherence}).
In fact, this provides an easy criterion for the existence of (unique)
preferred answer sets.
%
\begin{corollary}\label{thm:stratified:perfect:incoherence:W}
  Let $X^\star$ be the perfect model of stratified logic program $\Pi$
  and 
  let $<_s$ be an order induced by some stratification of $\Pi$.
  Let $(\Pi,<)$ be an ordered logic program such that $<\;\subseteq\; <_s$.
  
  Then , $(\Pi,<)$ has the unique preferred answer set $X^\star$.
\end{corollary}

\paragraph{Implementation through compilation.}

A translation of ordered logic programs to standard programs is developed
in~\cite{descto00c}.
Although the employed strategy (cf.\ Section~\ref{sec:delgrande:schaub:tompits})
differs from the one put forward in the previous section,
it turns out that the computation of preferred answer sets
can be accomplished by means of this translation technique
in a rather straightforward way.
In the framework of~\cite{descto00c},
preferences are expressed within the program via a predicate symbol $\prec$.
A logic program over a propositional language
$\mathcal{L}$ is said to be \emph{dynamically} ordered iff
$\mathcal{L}$ contains the following pairwise disjoint categories:
(i) a set $\nameset$ of terms serving as \emph{names} for rules;
(ii) a set $\Atm$ of atoms; and
(iii) a set $\Atm_{\prec}$ of \emph{preference atoms} $\PREC{s}{t}$,
  where $s,t\in\nameset$ are names.
For a program $\Pi$, we need a bijective function $\namef{\cdot}$
assigning a name $\namef{r}\in\nameset$ to each rule $r\in\Pi$ .
We sometimes write $\name{r}$ instead of $\namef{r}$.
An atom $\PREC{\name{r}}{\name{r'}}\in \Atm_{\prec}$ amounts to
asserting that $r<r'$ holds.
A (statically) ordered program $(\Pi,<)$ can thus be captured by programs
containing preference atoms only among their facts;
it is then expressed by the program
\(
\Pi\cup\{(\PREC{\name{r}}{\name{r'}})\LPif{}\mid\PRECM{r}{r'}\}
\).

Given $r<r'$,
one wants to ensure that $r'$ is considered before $r$
(cf.\ Condition~II in Definition~\ref{def:fixpoint:operator:W}).
For this purpose, one needs to be able to detect when a rule has been applied
or when a rule is defeated.
For detecting blockage, a new atom $\blocked{\name{r}}$ is introduced 
for each $r$ in $\Pi$.
Similarly, an atom $\applied{\name{r}}$ is introduced to indicate that a rule
has been applied.
For controlling application of rule $r$ the atom $\ok{\name{r}}$ is introduced.
Informally, one concludes that it is $\oksym$ to apply a rule just if it is
$\oksym$ with respect to every $<$-greater rule;
for such a $<$-greater rule $r'$, this will be the case just when $r'$ is
known to be blocked or applied.

More formally,
given a dynamically ordered program $\Pi$ over $\mathcal{L}$,
let $\mathcal{L}^{+}$ be the language obtained from $\mathcal{L}$ by adding,
for each $r,r'\in\Pi$, new pairwise distinct propositional atoms
\applied{\name{r}}, \blocked{\name{r}}, \ok{\name{r}}, and
\oko{\name{r}}{\name{r'}}.
Then, the translation $\Transo$ maps an ordered program $\Pi$ over
$\mathcal{L}$ into a standard program $\Trans{\Pi}$ over $\mathcal{L}^{+}$
in the following way.
%
\begin{definition}\label{def:compilation:W}
Let $\Pi = \{r_1,\dots, r_k\}$ be a dynamically ordered logic program over
$\mathcal{L}$.

Then,
the logic program $\Trans{\Pi}$ over $\mathcal{L}^{+}$ is defined as
\(
\Trans{\Pi}
=
\mbox{$\bigcup_{r\in\Pi}$}\tau(r)
\ ,
\)
where $\tau(r)$ consists of the following rules,
for
$L^{+}\in\pbody{r}$,
$L^{-}\in\nbody{r}$,
and
$r',r'' \in \Pi$~:
\[
\begin{array}{rcrcl}
  \RULE{\ap{1}{r}}
       {\head{r}}
       {\applied{\name{r}} }
  \\
  \RULE{\ap{2}{r}}
       {\applied{\name{r}}}
       {\ok{\name{r}},\body{r}}
  \\
  \RULE{\bl{1}{r}{L^{+}}}
       {\blocked{\name{r}}}
       {\ok{\name{r}}, \naf L^{+}}
  \\
  \RULE{\bl{2}{r}{L^{-}}}
       {\blocked{\name{r}}}
       {\ok{\name{r}},L^{-}}
  \\[1ex]
  \RULE{\cokt{1}{r}}
       {\ok{\name{r}}}
       {\oko{\name{r}}{\name{r_1}},\dots,\oko{\name{r}}{\name{r_k}}}
  \\
  \RULE{\cok{2}{r}{r'}}
       {\oko{\name{r}}{\name{r'}}}
       {\naf(\PREC{\name{r}}{\name{r'}})}
  \\
  \RULE{\cok{3}{r}{r'}}
       {\oko{\name{r}}{\name{r'}}}
       {(\PREC{\name{r}}{\name{r'}}),\applied{\name{r'}}}
  \\
  \RULE{\cok{4}{r}{r'}}
       {\oko{\name{r}}{\name{r'}}}
       {(\PREC{\name{r}}{\name{r'}}),\blocked{\name{r'}}}
  \\
  \RULE{\cok{5}{r}{r'}}
       {\oko{\name{r}}{\name{r'}}}
       {(\PREC{\name{r}}{\name{r'}}),\head{r'}}
  \\[1ex]
  \RULE{t(r,r',r'')}
       {\PREC{\name{r}}{\name{r''}}}
       {\PREC{\name{r}}{\name{r'}},\PREC{\name{r'}}{\name{r''}}}
  \\
  \RULE{as(r,r')}
       {{\neg(\PREC{\name{r'}}{\name{r}})}}
       {\PREC{\name{r}}{\name{r'}}}
\end{array}
\]
\end{definition}
%
We write $\Trans{\Pi,\PRECMo}$ rather than $\Trans{\Pi'}$, whenever
$\Pi'$ is the dynamically ordered program capturing $(\Pi,\PRECMo)$.
The first four rules of $\tau(r)$ express applicability 
and blocking conditions of the original rules.
For each rule $r\in\Pi$, we obtain two rules, \ap{1}{r} and \ap{2}{r}, along
with $n$ rules of the form \bl{1}{r}{L^{+}} and $m$ rules of the form \bl{2}{r}{L^{-}},
where $n$ and $m$ are the numbers of the
literals in $\pbody{r}$ and $\nbody{r}$, respectively.
The second group of rules encodes the strategy for handling preferences.
The first of these rules, \cokt{1}{r}, ``quantifies'' over the rules in $\Pi$.
This is necessary when dealing with dynamic preferences since preferences may
vary depending on the corresponding answer set.
The four rules \cok{i}{r}{r'} for $i=2\text{..}5$ specify the pairwise dependency of
rules in view of the given preference ordering:
For any pair of rules $r$, $r'$,
we derive $\oko{\name{r}}{\name{r'}}$ whenever $\name{r}\prec\name{r'}$ fails 
to hold,
or otherwise
whenever either $\applied{\name{r'}}$ or $\blocked{\name{r'}}$ is true, 
or whenever \head{r'} has already been derived.
This allows us to derive $\ok{\name{r}}$, indicating that $r$ may potentially
be applied whenever we have for all $r'$ with $\name{r}\prec\name{r'}$ that
$r'$ has been applied or cannot be applied.

It is instructive to observe how close this specification of \ok{\cdot} and
\oko{\cdot}{\cdot} is to Condition~II in Definition~\ref{def:Tp:W}.
In fact, given a fixed $r\in\Pi$, Condition~II can be read as follows.
{\itshape
\[
\begin{array}{rl}
  \mathit{II}. & \text{for every }r'\in\Pi\text{ with } r< r'\text{ either}
                 \\ 
               & (a)\ r'
                    \text{ is not active wrt } (Y,X)
                    \text{ or}
                 \\
               & (b)\ \head{r'}\in X
\end{array}
\]}%
\noindent
The quantification over all rules $r'\in\Pi$ with $r< r'$ is accomplished by
means of \cokt{1}{r} (along with \cok{2}{r}{r'}).
By definition, $r'$ is not active wrt $(Y,X)$\footnote{Recall that $X$ is
  supposed to contain the set of conclusions that have been derived so far,
  while $Y$ provides the putative answer set.} 
if either $\pbody{r}\not\subseteq Y$ or $\nbody{r}\cap X\neq\emptyset$, both of
which are detected by rule \cok{4}{r}{r'}.
The condition $\head{r'}\in X$ is reflected by \cok{3}{r}{r'} and \cok{5}{r}{r'}.
While the former captures the case where \head{r'} was supplied by $r'$
itself,\footnote{Strictly speaking rule \cok{3}{r}{r'} is subsumed by
  \cok{5}{r}{r'}; nonetheless we keep both for conceptual clarity in view of
  similar translations presented in Section~\ref{sec:fp:others}.}
the latter accounts additionally for the case where \head{r'} was supplied by
another rule than $r'$.

The next result shows that translation $\Transo$ is a realization of operator
$\mathcal{C}$.
%
\begin{theorem}\label{thm:CP:C:W}
  Let $(\Pi,<)$ be an ordered logic program over $\mathcal{L}$
  and
  let $X\subseteq\{\head{r}\mid r\in\Pi\}$
  be a consistent set of literals.
  Then,
  there is some set of literals $Y$ over $\mathcal{L^{+}}$
  where $X=Y\cap\mathcal{L}$ such that
  \(
  \CPXdefault= C_{\Trans{\Pi,<}}(Y)\cap\mathcal{L}
  \).
\end{theorem}
%
Note that the fixpoints of \CPOdefault\ constitute a special case the previous
theorem.
%
\begin{theorem}\label{thm:CP:T:W}
  Let $(\Pi,<)$ be an ordered logic program over $\mathcal{L}$
  and let $X$ and $Y$ be consistent sets of literals.   
  Then, we have that
  \begin{enumerate}
  \item if\/
    \(
    \CPXdefault= X
    \),
    then there is an answer set $Y$ of\/ $\Trans{\Pi,<}$
    such that
    \(
    X=Y\cap\mathcal{L}
    \);
  \item if\/ $Y$ is an answer set of\/ $\Trans{\Pi,<}$,
    then 
    \(
    \CPX{\Pi}{<}{Y\cap\mathcal{L}}={Y\cap\mathcal{L}}
    \).
  \end{enumerate}
\end{theorem}
%


\section{Other strategies (and characterizations)}
\label{sec:fp:others}

We now show how the approaches of Delgrande et al.~\shortcite{descto00c} and
Brewka/Eiter~\shortcite{breeit99a,breeit98a} can be captured within our
framework.
Also, we take up a complementary characterization provided
in~\cite{descto00c} in order to obtain another insightful perspective on
the three approaches.
For clarity, we add the letter ``\WZL'' to all concepts from Section~\ref{sec:fp}.
Accordingly we add ``\DST'' and ``\BE'', respectively, when dealing with
the two aforementioned approaches.

\paragraph{Characterizing \DST-preference.}
\label{sec:delgrande:schaub:tompits}
%
%
In~\cite{descto00c},
the selection of preferred answer sets is characterized in terms of the
underlying set of generating rules:
The set $\GR{\Pi}{X}$ of all \emph{generating rules} of a(n answer) set $X$ of
literals from program $\Pi$ is given by
\[
\GR{\Pi}{X}
=
\{r\in\Pi\mid\pbody{r}\subseteq X\text{ and }\nbody{r}\cap X=\emptyset\}
\ .
\]

The property distinguishing preferred answer sets from ordinary ones is
referred to as \emph{order preservation} and defined in the following way.
%
\begin{definition}\label{def:order:preservation:D}
  Let $(\Pi,<)$ be an ordered program
  and
  let $X$ be an answer set of $\Pi$.
  
  Then, $X$ is called  $<^{\DST}$-preserving,
  if there exists an enumeration
  \(
  \langle r_i\rangle_{i\in I}
  \)
  of\/ $\GR{\Pi}{X}$
  such that for every $i,j\in I$ we have that:
  \begin{enumerate}
  \item $\pbody{r_i}\subseteq\{\head{r_j}\mid j<i\}$;
    \ and
  \item if $r_i<r_j$, then $j<i$;
    \ and
  \item if
    $\PRECM{r_i}{r'}$
    and 
    \(
    r'\in {\Pi\setminus\GR{\Pi}{X}},
    \)
    then
    \begin{enumerate}
    \item 
      $\pbody{r'}\not\subseteq X$
      or
    \item 
      $\nbody{r'}\cap\{\head{r_j}\mid j<i\}\neq\emptyset$.
    \end{enumerate}
  \end{enumerate}
\end{definition}
%
We often refer to $<^{\DST}$-preserving answer sets as \DST-preferred answer
sets.

Condition~1 makes the property of \emph{groundedness}%
\footnote{This term is borrowed from the literature on default logic
  (cf.~\cite{konolige88a,schwind90}).}
explicit.
Although any standard answer set enjoys this property,
we will see that its interaction with preferences varies with the
strategy.
Condition~2 stipulates that $\langle r_i\rangle_{i\in I}$ is \emph{compatible}
with $<$, a property invariant to all of the considered approaches.
Lastly,
Condition~3 is comparable with Condition~II in Definition~\ref{def:Tp:W};
it guarantees that rules can never be blocked by lower-ranked rules.

Roughly speaking, an order preserving enumeration of the set of generating
rules reflects the sequence of successive rule applications leading to some
preferred answer set.
For instance, the preferred answer set $X=\{p,b,\neg f,w\}$ of
Example~(\ref{ex:bh}) can be generated by the two order preserving sequences
\(
\langle
r_5,r_4,r_1,r_2
\rangle
\)
and
\(
\langle
r_5,r_1,r_4,r_2
\rangle
\).
Intuitively, both enumerations are order preserving since they reflect the
fact that $r_1$ is treated before $r_2$.
\footnote{Note that both enumerations are compatible with the iteration through
  $\TPi{i}{\Pi_{\ref{ex:bh}}}{<}{X}{\emptyset}$ for $i=0\text{..}4$.}
Although there is another grounded enumeration generating $X$, namely
\(
\langle
r_5,r_4,r_2,r_1
\rangle
\), it is not order preserving since it violates Condition~2.
The same applies to the only grounded enumeration
\(
\langle
r_5,r_4,r_2,r_3
\rangle
\)
that allows to generate the second standard answer set of $\Pi_{\ref{ex:bh}}$;
it violates Condition~3b.
Consequently, $X$ is the only $<^{\DST}$-preserving answer set
of~$(\Pi_{\ref{ex:bh}},<)$.

%
We are now ready to provide a fixpoint definition for \DST-preference.
For this purpose,
we assume a bijective mapping \rul{\cdot} among rule heads and rules, that is,
\(
\rul{\head{r}}=r
\);
accordingly,
\(
\rul{\{\head{r}\mid r\in R\}}=R
\).
Such mappings can be defined in a bijective way by distinguishing different
occurrences of literals.
%
\begin{definition}\label{def:Tp:D}
Let $(\Pi,<)$ be an ordered logic program and let $X$ and $Y$ be sets of literals.

We define the set of 
immediate \DST-consequences of $X$ with respect to $(\Pi,<)$ and $Y$
as
\[
\TPD{\Pi}{<}{Y}{X}
\quad=\quad
\left\{
  \head{r}\left|\;
    \begin{array}{rl}
      \mathit{I}. & r\in\Pi
                    \text{ is active wrt } (X,Y)
                    \text{ and}
      \\
      \mathit{II}. & \text{there is no rule }r'\in\Pi
                     \text{ with } r< r'
                     \\ 
                   & \text{such that}
                     \\
                   & (a)\ r'
                        \text{ is active wrt } (Y,X)
                        \text{ and}
                        \\
                   & (b)\ r'\not\in\rul{X}
    \end{array}
  \right\}\right.
\]

if $X$ is consistent, and $\TPDX{\Pi}{<}{Y}{X}=\Lit$ otherwise.
\end{definition}
%
The distinguishing feature between this definition and
Definition~\ref{def:Tp:W} manifests itself in IIb.
While \DST-preference requires that a higher-ranked rule has effectively
applied,
\WZL-preference contents itself with the presence of the head of the rule,
no matter whether this was supplied by the rule itself.

Defining iterated applications of $\TPDo{\Pi}{<}{Y}$ in analogy to those of
$\TPo{\Pi}{<}{Y}$,
we may capture \DST-preference by means of a fixpoint operator in the
following way.
%
\begin{definition}\label{def:fixpoint:operator:D}
\renewcommand{\TPo}[3]{\ensuremath{(\mathcal{T}^{\DST})_{(#1,#2),#3}}}
\renewcommand{\TPDiO}[4]{\TPo{#2}{#3}{#4}^{#1}}
Let $(\Pi,<)$ be an ordered logic program and let $X$ be a set of literals.

We define
\(
\CPDXdefault=\bigcup_{i\geq 0}\TPDi{i}{\Pi}{<}{X}{\emptyset}
\).
\end{definition}
%
A similar elaboration of \CPDOdefault\ as done with \CPWOdefault\ in
Section~\ref{sec:fixpoint:properties} yields identical formal properties; in
particular, \CPDOdefault\ also enjoys anti-monotonicity.

The aforementioned difference is nicely illustrated by extending the
programs in~(\ref{ex:incoherence}) by rule $a\LPif{}$,
yielding $({\Pi_{\ref{ex:DvsW}}},<)$ and $({\Pi'_{\ref{ex:DvsW}}},<')$,
respectively:
\begin{equation}\label{ex:DvsW}
  \begin{array}[c]{rcrcl}
    r_1 & = & a & \leftarrow &      b \\
    r_2 & = & b & \leftarrow &        \\
    r_3 & = & a & \leftarrow &        \\[1ex]
    &\multicolumn{3}{l}{r_2< r_1}
  \end{array}
  \qquad\qquad\qquad
  \begin{array}[c]{rcrcl}
    r'_1 & = & a & \leftarrow & \naf b \\
    r'_2 & = & b & \leftarrow &        \\
    r'_3 & = & a & \leftarrow &        \\[1ex]
    &\multicolumn{3}{l}{r'_2<' r'_1}
  \end{array}
\end{equation}
While in both cases the single standard answer set is \WZL-preferred,
neither of them is \DST-preferred.
Let us illustrate this in terms of the iterated applications of
$\TPWo{\Pi_{\ref{ex:DvsW}}}{<}{X}$ and $\TPDo{\Pi_{\ref{ex:DvsW}}}{<}{X}$,
where $X=\{a,b\}$ is the standard answer set of ${\Pi_{\ref{ex:DvsW}}}$:
At first, both operators allow for applying rule $a\LPif{}$, resulting in $\{a\}$.
As with $\TPWo{\Pi_{\ref{ex:incoherence}}}{<}{X}$ in~(\ref{ex:incoherence}),
however,
operator $\TPDo{\Pi_{\ref{ex:DvsW}}}{<}{X}$ does not allow for applying $r_2$
at the next stage, unless $r_1$ is inactive.
This requirement is now dropped by $\TPWo{\Pi_{\ref{ex:DvsW}}}{<}{X}$,
since the head of $r_1$ has already been derived through $r_3$.
In such a case, the original preference is ignored, which enables the
application of $r_2$.
In this way, we obtain the \WZL-preferred answer set  $X=\{a,b\}$.
The analogous behavior is observed on  $({\Pi'_{\ref{ex:DvsW}}},<')$.

As \WZL-preferred answer sets, \DST-preferred ones coincide with the
perfect model on stratified programs.
%
\begin{theorem}\label{thm:stratified:perfect:D}
  Let $X^\star$ be the perfect model of stratified logic program $\Pi$
  and 
  let $<_s$ be an order induced by some stratification of $\Pi$.
  Then , $(\Pi,<_s)$ has the unique \DST-preferred answer set $X^\star$.
\end{theorem}

The subtle difference between \DST- and \WZL-preference is also reflected in
the resulting compilation.
%
Given the same prerequisites as in Definition~\ref{def:compilation:W},
the logic program $\TransD{\Pi}$ over $\mathcal{L}^{+}$ is defined as
\(
\TransD{\Pi}
=
\TransW{\Pi}\setminus\{\cok{5}{r}{r'}\mid r,r'\in\Pi\}
\).
%
Hence, in terms of this compilation technique,
the distinguishing feature between \DST- and \WZL-preference manifests itself
in the usage of rule
\(
\cok{5}{r}{r'}:
{\oko{\name{r}}{\name{r'}}}\LPif{}{(\PREC{\name{r}}{\name{r'}}),\head{r'}}
\).
While \WZL-preference allows for suspending a preference whenever the head of
the preferred rule was derived,
\DST-preference stipulates the application of the preferred rule itself.
This is reflected by the fact that the translation $\TransDo$ merely uses
rule
\(
{\cok{3}{r}{r'}}:
{\oko{\name{r}}{\name{r'}}}\LPif{}{(\PREC{\name{r}}{\name{r'}}),\applied{\name{r'}}}
\)
to enforce that the preferred rule itself has been applied.
This demonstrates once more how closely the compilation technique follows the
specification given in the fixpoint operation.

As shown in \cite{descto00c},
a set of literals
$X$ is a $<^{\DST}$-preserving answer set of a program $\Pi$
iff
$X=Y\cap\mathcal{L}$ for some answer set $Y$ of\/ $\TransD{\Pi,\PRECMo}$.
This result naturally extends to the fixpoint operator
$\mathcal{C}^{\DST}_{(\Pi,<)}$, as shown in the following result.
%
\begin{theorem}\label{thm:results:D}
  Let $(\Pi,<)$ be an ordered logic program over $\mathcal{L}$
  and let $X$ be a consistent set of literals.
  Then, the following propositions are equivalent.
  \begin{enumerate}
  \item 
    \(
    \mathcal{C}^{\DST}_{(\Pi,<)}(X)= X
    \);
  \item 
    \(
    X=Y\cap\mathcal{L}
    \)
    for some answer set $Y$ of\/ $\TransD{\Pi,<}$;
  \item $X$ is a $<^{\DST}$-preserving answer set of\/ $\Pi$.
  \end{enumerate}
\end{theorem}
%
While the last result dealt with effective answer sets, the next one shows
that applying $\mathcal{C}^{\DST}_{(\Pi,<)}$ is equivalent to the application
of $C_{\Pi'}$ to the translated program $\Pi'={\TransD{\Pi,<}}$ .
%
\begin{theorem}\label{thm:CP:C:D}
  Let $(\Pi,<)$ be an ordered logic program over $\mathcal{L}$
  and
  let $X\subseteq\{\head{r}\mid r\in\Pi\}$
  be a consistent set of literals.
  Then,
  there is some set of literals $Y$ over $\mathcal{L^{+}}$
  where $X=Y\cap\mathcal{L}$ such that
  \(
  \mathcal{C}^{\DST}_{(\Pi,<)}(X)= C_{\TransD{\Pi,<}}(Y)\cap\mathcal{L}
  \).
\end{theorem}

\paragraph{Characterizing \WZL-preference (alternatively).}
\label{sec:fp:order:preservation}

We now briefly elaborate upon a characterization of \WZL-preference in
terms of order preservation.
This is interesting because order preservation provides an alternative
perspective on the formation of answer sets.
In contrast to the previous fixpoint characterizations,
order preservation furnishes an account of preferred answer sets in terms of
the underlying generating rules.
While an immediate consequence operator provides a rather rule-centered and
thus local characterization,
order preservation gives a more global and less procedural view on an entire
construction.
In particular, the underlying sequence nicely reflects the interaction of its
properties.
In fact, we see below that different approaches distinguish themselves by a
different degree of interaction between groundedness and preferences.
%
\begin{definition}\label{def:order:preservation:W}
  Let $(\Pi,<)$ be an ordered program
  and
  let $X$ be an answer set of\/ $\Pi$.
  
  Then, $X$ is called  $<^{\WZL}$-preserving,
  if there exists an enumeration
  \(
  \langle r_i\rangle_{i\in I}
  \)
  of\/ $\GR{\Pi}{X}$
  such that for every $i,j\in I$ we have that:
  \begin{enumerate}
  \item 
    \begin{enumerate}
    \item $\pbody{r_i}\subseteq\{\head{r_j}\mid j<i\}$
      or
    \item $\,\head{r_i}\;\in\,\{\head{r_j}\mid j<i\}$;
      \ and
    \end{enumerate}
  \item if $r_i<r_j$, then $j<i$;
    \ and
  \item if
    $\PRECM{r_i}{r'}$
    and 
    \(
    r'\in {\Pi\setminus\GR{\Pi}{X}},
    \)
    then
    \begin{enumerate}
    \item 
      $\pbody{r'}\not\subseteq X$
      or
    \item 
      $\nbody{r'}\cap\{\head{r_j}\mid j<i\}\neq\emptyset$
      or
    \item
      $\,\head{r'}\,\in \{\head{r_j}\mid j<i\}$.
    \end{enumerate}
  \end{enumerate}
\end{definition}
%
The primary difference between this concept of order preservation and the one
for \DST-preference is clearly the weaker notion of groundedness.
While \DST-preference makes no compromise when enforcing rule dependencies
induced by preference, \WZL-preference ``smoothes'' their integration with
those induced by groundedness and defeat relationships:
First, regarding rules in \GR{\Pi}{X} (via Condition~1b)
and
second concerning rules in ${\Pi\setminus\GR{\Pi}{X}}$ (via Condition~3c).
The rest of the definition is identical to Definition~\ref{def:order:preservation:D}.

This ``smoothed'' integration of preferences with groundedness and defeat
dependencies is nicely illustrated by programs $(\Pi_{\ref{ex:DvsW}},<)$ and
$(\Pi_{\ref{ex:DvsW}}',<)$.
Regarding $\Pi_{\ref{ex:DvsW}}$,
we observe that there is no enumeration of\/ $\GR{\Pi}{X}$ satisfying both
Condition~1a and 2.
Rather it is Condition~1b that weakens the interaction between both conditions
by tolerating enumeration
\(
\langle r_3,r_2,r_1\rangle
\).
A similar observation can be made regarding $\Pi'_{\ref{ex:DvsW}}$,
where, in contrast to $\Pi_{\ref{ex:DvsW}}$, the preferred rule $r'_1$ does
not belong to \GR{\Pi}{X}.
We observe that there is no enumeration of\/ $\GR{\Pi}{X}$ satisfying both
Condition~2 and~3a/b.
Now, it is Condition~3c that weakens the interaction between both conditions
by tolerating enumeration
\(
\langle r'_3,r'_2\rangle
\).
In fact, the two examples show that both Condition~1b as well as~3c function
as exceptions to conditions~1a and~3a/b, respectively.
In this way,
\WZL-preference imposes the same requirements as \DST-preference,
\emph{unless} the head of the rule in focus has already been derived by other
means.

Finally, we have the following summarizing result.
%
\begin{theorem}\label{thm:results:W}
  Let $(\Pi,<)$ be an ordered logic program over $\mathcal{L}$
  and let $X$ be a consistent set of literals.   
  Then, the following propositions are equivalent.
  \begin{enumerate}
  \item 
    \(
    \CPWXdefault= X
    \);
  \item 
    \(
    X=Y\cap\mathcal{L}
    \)
    for some answer set $Y$ of\/ $\TransW{\Pi,<}$;
  \item $X$ is a $<^{\WZL}$-preserving answer set of\/ $\Pi$.
  \end{enumerate}
\end{theorem}

\paragraph{Characterizing \BE-preference.}
\label{sec:breiter}
Another approach to preference is proposed in~\cite{breeit99a}.
This approach differs in two ways from the previous ones.
First, the construction of answer sets is separated from verifying preferences.
Interestingly, this verification is done on the basis of the prerequisite-free
program obtained from the original one by ``evaluating'' \pbody{r} for each
rule $r$ wrt the separately  constructed (standard) answer set.
Second, rules that may lead to counter-intuitive results are explicitly
removed.
This is spelled out in~\cite{breeit98a}, where the following filter is defined:
\begin{equation}
  \label{eq:brewka:eiter:filter}
  \mathcal{E}_X(\Pi)
  =
  \Pi\setminus\{r\in\Pi\mid\head{r}\in X,\nbody{r}\cap X\neq\emptyset\}
\end{equation}
Accordingly, we define
\(
\mathcal{E}_X(\Pi,<)
=
(\mathcal{E}_X(\Pi),<\cap\;{(\mathcal{E}_X(\Pi)\times\mathcal{E}_X(\Pi))}\,)
\).

%
We begin with a formal account of \BE-preferred answer sets.
In this approach, partially ordered programs are reduced to totally ordered
ones:
A \emph{fully ordered logic program} is an ordered logic program
$(\Pi,\ll)$ where $\ll$ is a total ordering.  
The case of arbitrarily ordered programs is reduced to this restricted case:
%
Let $(\Pi,\PRECMo)$ be an ordered logic program 
and let
$X$ be a set of literals. 
Then,
$X$ is a \BE-preferred answer set of $(\Pi,\PRECMo)$ 
iff
$X$ is a \BE-preferred answer set of some fully ordered logic program
$(\Pi,\ll)$ such that ${\PRECMo}\subseteq{\ll}$.

The construction of \BE-preferred answer sets relies on an operator,
defined for prerequisite-free programs, comprising only rules $r$ with
$\pbody{r}=\emptyset$.
%
\begin{definition}\label{def:be:operator}
  Let $(\Pi,\ll)$ be a fully ordered prerequisite-free logic program,
  let
  \(
  \langle r_i\rangle_{i\in I}
  \)
  be an enumeration of $\Pi$ according to $\ll$, 
  and 
  let $X$ be a set of literals.
  Then,
  $\mathcal{B}_{(\Pi,\ll)}(X)$ is the smallest logically closed set of
  literals containing
  \(
  \bigcup_{i\in I} X_i
  \),
  where $X_j=\emptyset$ for $j\not\in I$ and
  \[
  X_i = 
  \left\{
    \begin{array}{ll}
      X_{i-1}
      &
      \text{ if }
      \nbody{r_i}\cap X_{i-1}\neq\emptyset
      \\
      X_{i-1}\cup\{\head{r_i}\}
      &
      \text{ otherwise.}
  \end{array}
  \right.
  \]
\end{definition}
%
This construction is unique insofar that for any such program
$(\Pi,\nolinebreak{\ll)}$,
there is at most one standard answer set $X$ of $\Pi$ such that 
\(
\mathcal{B}_{\mathcal{E}_X(\Pi,\ll)}(X)=X
\).
Accordingly, this set is used for defining the \emph{\BE-preferred answer set}
of a prerequisite-free logic program:
%
\begin{definition}\label{def:be:extension:supernormal}
  Let $(\Pi,\ll)$ be a fully ordered prerequisite-free logic program
  and let $X$ be a set of literals.
  Then,
  $X$ is the \BE-preferred answer set of $(\Pi,\ll)$
  iff\/
  \(
  \mathcal{B}_{\mathcal{E}_X(\Pi,\ll)}(X)=X
  \).
\end{definition}

The reduction of $(\Pi,\ll)$ to ${\mathcal{E}_X(\Pi,\ll)}$ removes from the
above construction all rules whose heads are in $X$ but which are defeated by
$X$.
This is illustrated in~\cite{breeit98a} through the following example:
\begin{equation}\label{eq:three}
  \begin{array}[t]{rcrcl}
    r_1 & = & a      &\LPif& \naf b,
    \\
    r_2 & = & \neg a &\LPif& \naf a,
  \end{array}
  \begin{array}[t]{rcrcl}
    r_3 & = & a      &\LPif& \naf\neg a,
    \\
    r_4 & = & b      &\LPif& \naf\neg b,
  \end{array}
  \quad
  \{\PRECM{r_j}{r_i}\mid i<j\}
  \ .
\end{equation}
Program $\Pi_{\ref{eq:three}}=\{r_1,\dots,r_4\}$ has two answer sets,
$\{a,b\}$ and $\{\neg a,b\}$.
The application of operator $\mathcal{B}$ relies on sequence
\(
\langle r_1,r_2,r_3,r_4 \rangle
\).
Now, consider the processes induced by
\(
\mathcal{B}_{\mathcal{E}_X(\Pi_{\ref{eq:three}},<)}(X)
\)
and
\(
\mathcal{B}_{(\Pi_{\ref{eq:three}},<)}(X)
\)
for $X=\{a,b\}$, respectively:
\[
\begin{array}{rp{3mm}llll}
  \mathcal{B}_{\mathcal{E}_X(\Pi_{\ref{eq:three}},<)}(X):&&
  X_1=\{ \} & X_2=\{\neg a\} & X_3=\{\neg a\} & X_4=\{\neg a,b\} 
  \\
  \mathcal{B}_{(\Pi_{\ref{eq:three}},<)}(X):&&
  X'_1=\{a\} &X'_2=\{     a\} &X'_3=\{     a\} &X'_4=\{     a,b\} 
\end{array}
\]
Thus, without filtering by ${\mathcal{E}_X}$, we get $\{a,b\}$ as a
\BE-preferred answer set.
As argued in~\cite{breeit98a}, such an answer set does not preserve priorities
because $r_2$ is defeated in $\{a,b\}$ by applying a rule which is less
preferred than $r_2$, namely $r_3$.
The above program has therefore no \BE-preferred answer set.

The next definition accounts for the general case by reducing it to the
prerequisite-free one.
For checking whether an answer set $X$ is \BE-preferred, 
the prerequisites of the rules are evaluated wrt $X$.
For this purpose, we define
\(
\reductBE{r}=\head{r}\LPif\nbody{r}
\)
for a rule $r$.
%
\begin{definition}\label{def:be:answer:set}
  Let $(\Pi,\ll)$ be a fully ordered logic program
  and
  $X$ a set of literals.

  The logic program $(\Pi_X,\ll_X)$ is obtained from $(\Pi,\ll)$ as follows:
  \begin{enumerate}
  \item\label{l:one:b}
    $\Pi_X=\{\reductBE{r}\mid r\in\Pi \text{ and } \pbody{r}\subseteq X\}$;
  \item\label{l:two}
  for any 
  $r'_1, r'_2 \in \Pi_X$, 
  $r'_1 \ll_X r'_2$
  iff $r_1 \ll r_2$ where
  \(
  r_i=\max_\ll\{r\in \Pi\mid\reductBE{r}=r'_i\}
  \).
\end{enumerate}
\end{definition}
%
In other words, $\Pi_X$ is obtained from $\Pi$ by
first eliminating every rule $r\in\Pi$ such that $\pbody{r}\not\subseteq X$,
and
then substituting all remaining rules $r$ by \reductBE{r}.

In general, \BE-preferred answer sets are then defined as follows.
%
\begin{definition}\label{def:be:extension:general}
  Let $(\Pi,\ll)$ be a fully ordered logic program
  and
  $X$ a set of literals.

  Then, $X$ is a \BE-preferred answer set of $(\Pi,\ll)$, 
  if
  \begin{enumerate}
  \item $X$ is a (standard) answer set of $\Pi$, and
  \item $X$ is a \BE-preferred answer set of $(\Pi_X,\ll_X)$.
  \end{enumerate}
\end{definition}

The distinguishing example of this approach is given by program
$(\Pi_{\ref{eq:five:one}},<)$:
\begin{equation}\label{eq:five:one}
  \begin{array}[t]{rcrcl}
    r_1 & = &      b & \LPif & a, \naf\neg b
    \\
    r_2 & = & \neg b & \LPif & \naf     b
    \\ 
    r_3 & = &      a & \LPif & \naf\neg a
  \end{array}
  \qquad\text{ with }\qquad
  \{\PRECM{r_j}{r_i}\mid i<j\}
  \ .
\end{equation} 
Program $\Pi_{\ref{eq:five:one}}=\{r_1,r_2,r_3\}$ has two standard answer sets: 
\(
X_1=\{a,b\}
\)
and
\(
X_2=\{a,\neg b\}
\).
Both $(\Pi_{\ref{eq:five:one}})_{X_1}$ as well as $(\Pi_{\ref{eq:five:one}})_{X_2}$
turn $r_1$ into
\(
b \LPif \naf\neg b
\)
while leaving $r_2$ and $r_3$ unaffected.
Clearly,
\(
\mathcal{E}_{X_i}(\Pi_{\ref{eq:five:one}},<)=(\Pi_{\ref{eq:five:one}},<)
\)
for $i=1,2$.
Also, we obtain that
\(
\mathcal{B}_{(\Pi_{\ref{eq:five:one}},<)}(X_1) = X_1
\), 
that is, $X_1$ is a \BE-preferred answer set.
In contrast to this, $X_2$ is not \BE-preferred.
To to see this, observe that
\(
\mathcal{B}_{(\Pi_{\ref{eq:five:one}},<)}(X_2)=X_1\neq X_2
\).
That is, $\mathcal{B}_{(\Pi_{\ref{eq:five:one}},<)}(X_2)$ reproduces $X_1$
rather than~$X_2$.
In fact, while $X_1$ is the only \BE-preferred set, neither $X_1$ nor $X_2$ is
\WZL- or \DST-preferred (see below).

We note that \BE-preference disagrees with \WZL- and \DST-preference on
Example~(\ref{ex:bh}).
In fact,
both answer sets of program $(\Pi_{\ref{ex:bh}},<)$ are \BE-preferred,
while only $\{p, b, \neg f, w\}$ is \WZL- and \DST-preferred.
%
%
In order to shed some light on these differences, we start by providing a
fixpoint characterization of \BE-preference:
%
\begin{definition}\label{def:Tp:B}
Let $(\Pi,<)$ be an ordered logic program and let $X$ and $Y$ be sets of literals.

We define the set of 
immediate consequences of $X$ with respect to $(\Pi,<)$ and $Y$
as
\[
\TPB{\Pi}{<}{Y}{X}
\quad=\quad
\left\{
  \head{r}\left|\;
    \begin{array}{rl}
      \mathit{I}. & r\in\Pi
                    \text{ is active wrt } (Y,Y)
                    \text{ and}
      \\
      \mathit{II}. & \text{there is no rule }r'\in\Pi
                     \text{ with } r< r'
                     \\ 
                   & \text{such that}
                     \\
                   & (a)\ r'
                        \text{ is active wrt } (Y,X)
                        \text{ and}
                        \\
                   & (b)\ \head{r'}\not\in X
    \end{array}
  \right\}\right.
\]

if $X$ is consistent, and $\TPBX{\Pi}{<}{Y}{X}=\Lit$ otherwise.
\end{definition}
%
The difference between this operator%
\footnote{We have refrained from integrating~(\ref{eq:brewka:eiter:filter})
  in order to keep the fixpoint operator comparable to its predecessors.
  This is taken care of in Theorem~\ref{thm:results:B}.
  We note however that an integration of~(\ref{eq:brewka:eiter:filter}) would
  only affect Condition~II.}
and its predecessors manifests itself in Condition~I,
where activeness is tested wrt $(Y,Y)$ instead of $(X,Y)$,
as in Definition~\ref{def:Tp:W} and~\ref{def:compilation:W}.
In fact, in Example~(\ref{eq:five:one}) it is the (unprovability of the)
prerequisite $a$ of the highest-ranked rule $r_1$ that makes the construction
of \WZL- or \DST-preferred answer sets break down
(cf.\ Definition~\ref{def:Tp:W} and~\ref{def:compilation:W}).
This is avoided with \BE-preference because once answer set $\{a,b\}$ is
provided, preferences are enforced wrt the program obtained by replacing $r_1$
with $b\LPif\naf\neg b$.

With an analogous definition of iterated applications of \TPBXdefault\ as above,
we obtain the following characterization of \BE-preference:
%
\begin{definition}\label{def:fixpoint:operator:B}
\renewcommand{\TPo}[3]{\ensuremath{(\mathcal{T}^{\BE})_{(#1,#2),#3}}}
\renewcommand{\TPBiO}[4]{\TPo{#2}{#3}{#4}^{#1}}
Let $(\Pi,<)$ be an ordered logic program and let $X$ be a set of literals.

We define
\(
\CPBXdefault=\bigcup_{i\geq 0}\TPBi{i}{\Pi}{<}{X}{\emptyset}
\).
\end{definition}
%
Unlike above, \CPBOdefault\ is not anti-monotonic.
This is related to the fact that the ``answer set property'' of a set is
verified separately (cf.\ Definition~\ref{def:be:extension:general}).
We have the following result.
%
\begin{theorem}\label{thm:CP:B}
  Let $(\Pi,<)$ be an ordered logic program over $\mathcal{L}$
  and let $X$ be an answer set of $\Pi$.

  Then, we have that
  $X$ is \BE-preferred
  iff
  \(
  \mathcal{C}^{\BE}_{\mathcal{E}_X(\Pi,<)}(X)= X
  \).
\end{theorem}
%
As with \DST- and \WZL-preference, \BE-preference gives the perfect model
on stratified programs.
%
\begin{theorem}\label{thm:stratified:perfect:B}
  Let $X^\star$ be the perfect model of stratified logic program $\Pi$
  and 
  let $<_s$ be an order induced by some stratification of $\Pi$.
  Then , $(\Pi,<_s)$ has the unique \BE-preferred answer set $X^\star$.
\end{theorem}

%
Alternatively, \BE-preference can also be captured by appeal to order preservation:
%
\begin{definition}\label{def:order:preservation:B}
  Let $(\Pi,<)$ be an ordered program
  and
  let $X$ be an answer set of\/ $\Pi$.

  Then, $X$ is called  $<^{\BE}$-preserving,
  if there exists an enumeration
  \(
  \langle r_i\rangle_{i\in I}
  \)
  of\/ $\GR{\Pi}{X}$
  such that, for every $i,j\in I$, we have that:
  \begin{enumerate}
    \item if $r_i<r_j$, then $j<i$;
          \ and
    \item if
          $\PRECM{r_i}{r'}$
          and 
          \(
          r'\in {\Pi\setminus\GR{\Pi}{X}},
          \)
          then
          \begin{enumerate}
          \item 
            $\pbody{r'}\not\subseteq X$
            or
          \item 
            $\nbody{r'}\cap\{\head{r_j}\mid j<i\}\neq\emptyset$
            or
          \item
            $\head{r'}\in X$.
          \end{enumerate}
        \end{enumerate}
\end{definition}
%
This definition differs in two ways from its predecessors.
First, it drops any requirement on groundedness.
This corresponds to using $(Y,Y)$ instead of $(X,Y)$ in
Definition~\ref{def:Tp:B}.
Hence, groundedness is fully disconnected from order preservation.
For example, the \BE-preferred answer set $\{a,b\}$ of
$(\Pi_{\ref{eq:five:one}},<)$ is associated with the 
$<^{\BE}$-preserving sequence
\(
\langle r_1,r_2\rangle
\),
while the standard answer set is generated by the grounded sequence
\(
\langle r_2,r_1\rangle
\).
Second,
Condition~2c is more relaxed than in Definition~\ref{def:order:preservation:W}.
That is,
any rule $r'$ whose head is in $X$ (as opposed to $\{\head{r_j}\mid j<i\}$) is
taken as ``applied''.
Also, 
Condition~2c integrates the filter in~(\ref{eq:brewka:eiter:filter}).%
\footnote{Condition ${\nbody{r'}\cap X}\neq\emptyset$ in
  (\ref{eq:brewka:eiter:filter}) is obsolete because $r'\not\in\GR{\Pi}{X}$.}
For illustration, consider Example~(\ref{ex:DvsW}) extended by $r_3< r_2$:
\begin{equation}\label{ex:DWvsB}
  \begin{array}[t]{rcrcl}
    r_1 & = & a & \leftarrow & \naf b \\
    r_2 & = & b & \leftarrow &        \\
    r_3 & = & a & \leftarrow & 
  \end{array}
  \qquad\qquad
  r_3< r_2< r_1
\end{equation}
While this program has no \DST- or \WZL-preferred answer set,
it has a \BE-preferred one: $\{a,b\}$ generated by
\(
\langle r_2,r_3\rangle
\).
The critical rule $r_1$ is handled by 2c.
As a net result, Condition~2 is weaker than its counterpart in
Definition~\ref{def:order:preservation:W}.
We have the following summarizing result.
%
\begin{theorem}\label{thm:results:B}
  Let $(\Pi,<)$ be an ordered logic program over $\mathcal{L}$
  and let $X$ be a consistent answer set of\/ $\Pi$.
  Then, the following propositions are equivalent.
  \begin{enumerate}
  \item $X$ is \BE-preferred;
  \item 
    \(
    \mathcal{C}^{\BE}_{\mathcal{E}_X(\Pi,<)}(X)= X
    \);
  \item
    $X$ is a $<^{\BE}$-preserving answer set of\/ $\Pi$\/;
  \item
    \(
    X=Y\cap\mathcal{L}
    \)
    for some answer set $Y$ of\/ $\TransB{\Pi,<}$
    \\
    (where $\TransBo$ is defined in \cite{descto00d}).
  \end{enumerate}
\end{theorem}
%
Unlike theorems~\ref{thm:results:D} and~\ref{thm:results:W},
the last result stipulates that $X$ must be an answer set of $\Pi$.
This requirement can only be dropped in case 4,
while all other cases rely on this property.

\paragraph{Relationships.}
%
First of all, we observe that all three approaches treat the blockage of
(higher-ranked) rules in the same way.
That is, a rule $r'$ is found to be blocked if either its prerequisites in
\pbody{r'} are \emph{never} derivable or if some member of \nbody{r'} has been
derived by higher-ranked or unrelated rules.
This is reflected by the identity of conditions IIa and 2a/b in all three
approaches, respectively.
Although this is arguably a sensible strategy, it leads to the loss of
preferred answer sets on programs like~$(\Pi'_{\ref{ex:incoherence}},<')$.

The difference between \DST- and \WZL-preference can be directly read off
Definition~\ref{def:Tp:W} and~\ref{def:compilation:W}; it manifests itself in
Condition~IIb and leads to the following relationships.
%
\begin{theorem}\label{thm:D:W:eq}
  Let $(\Pi,<)$ be an ordered logic program 
  such that for $r,r'\in\Pi$ we have that
  $r\neq r'$ implies $\head{r}\neq\head{r'}$.
  Let $X$ be a set of literals.
  Then, 
  $X$ is a \DST-preferred answer set of  $(\Pi,<)$
  iff
  $X$ is a \WZL-preferred answer set of  $(\Pi,<)$.
\end{theorem}
%
The considered programs deny the suspension of preferences under
\WZL-preference, because all rule heads are derivable in a unique way.
We have the following general result.
%
\begin{theorem}\label{thm:D:W}
  Every \DST-preferred answer set is \WZL-preferred. 
\end{theorem}
%
Example~(\ref{ex:DvsW}) shows that the converse does not hold.

Interestingly,
a similar relationship is obtained between \WZL- and \BE-preference.
In fact, Definition~\ref{def:order:preservation:B} can be interpreted as a
weakening of Definition~\ref{def:order:preservation:W} by dropping
the condition on groundedness and weakening Condition~2 (via~2c).
We thus obtain the following result.
%
\begin{theorem}\label{thm:W:B}
  Every \WZL-preferred answer set is \BE-preferred. 
\end{theorem}
%
Example~(\ref{eq:five:one}) shows that the converse does not hold.

Let
\(
\mathcal{AS} (\Pi)=\{X\mid C_{\Pi}(X)=X\}
\)
and
\(
\mathcal{AS}_P(\Pi,<)=\{X\in\mathcal{AS} (\Pi)\mid X\text{ is }P\text{-preferred}\}
\)
for $P=\WZL{},\DST{},\BE$.
Then,
we obtain the following summarizing result.
%
\begin{theorem}\label{thm:D:W:B}
  Let $(\Pi,<)$ be an ordered logic program.
  Then, we have
  \[
  \mathcal{AS}_\DST(\Pi,<)
  \subseteq
  \mathcal{AS}_\WZL(\Pi,<)
  \subseteq
  \mathcal{AS}_\BE(\Pi,<)
  \subseteq
  \mathcal{AS}(\Pi)
  \]
\end{theorem}
%
This hierarchy is primarily induced by a decreasing interaction between
groundedness and preference.
While \DST-preference requires the full compatibility of both concepts,
this interaction is already weakened in \WZL-preference,
before it is fully abandoned in \BE-preference.
This is nicely reflected by the evolution of the condition on groundedness in
definitions~\ref{def:order:preservation:D},~\ref{def:order:preservation:W},
and~\ref{def:order:preservation:B}.
Notably, groundedness as such is not the ultimate distinguishing factor,
as demonstrated by the fact that prerequisite-free programs do not necessarily
lead to the same preferred answer sets,
as witnessed in~(\ref{ex:DvsW}) and~(\ref{ex:DWvsB}).
Rather it is the degree of integration of preferences within the standard
reasoning process that makes the difference.

Taking together theorems~\ref{thm:stratified:perfect:W:ii},
\ref{thm:stratified:perfect:D}, and \ref{thm:stratified:perfect:B},
we obtain the following result.
%
\begin{theorem}\label{thm:D:W:B:stratified}
  Let $X^\star$ be the perfect model of stratified logic program $\Pi$
  and 
  let $<_s$ be an order induced by some stratification of $\Pi$.
  Let $(\Pi,<)$ be an ordered logic program such that $<\;\subseteq\; <_s$.
  
  Then, we have
  \(
  \mathcal{AS}_\DST(\Pi,<)
  =
  \mathcal{AS}_\WZL(\Pi,<)
  =
  \mathcal{AS}_\BE(\Pi,<)
  =
  \mathcal{AS}(\Pi)
  =
  \{X^\star\}
  \).
\end{theorem}
%



\section{Discussion and related work}
\label{sec:discussion}

Up to now, we have been dealing with static preferences only.
In fact, all fixpoint characterizations are also amenable to dynamically
ordered programs, as introduced in Section~\ref{sec:delgrande:schaub:tompits}.
To see this,
consider Definition~\ref{def:Tp:W} along with a dynamically ordered program
$\Pi$ and sets of literals $X,Y$ over a language extended by preference atoms
$\Atm_{\prec}$.
Then, the corresponding preferred answer sets are definable by substituting
``$r< r'$'' by ``$(\PREC{r}{r'})\in Y$'' in
definitions~\ref{def:Tp:W},~\ref{def:Tp:D}, and~\ref{def:Tp:B}, respectively.
That is, instead of drawing preference information from the external order
$<$, we simply consult the initial context, expressed by $Y$.
In this way, the preferred answer sets of $\Pi$ can be given by the fixpoints
of an operator $\mathcal{C}_\Pi$.

Also, we have concentrated so far on preferred answer sets semantics that
amount to selection functions on the standard answer sets of the underlying
program.
Another strategy is advocated in~\cite{gelson97a}, where the preference $d_1<d_2$
\emph{``stops the application of default $d_2$ if defaults $d_1$ and $d_2$ are
  in conflict with each other and the default $d_1$ is
  applicable''}~\cite{gelson97a}. 
In contrast to \BE{}-, \DST{}-, and \WZL{}-preference this allows for
exclusively concluding $\neg p$ from program $(\{r_1,r_2\},<)$:
\[ 
  \begin{array}[t]{rcrcl}
    r_1 & = & p & \leftarrow &
  \end{array}
  \qquad
  \begin{array}[t]{rcrcl}
    r_2 & = & \neg p & \leftarrow &
  \end{array}
  \qquad\qquad
  r_1< r_2
\] 
This approach amounts to \BE{}-preference on certain ``hierarchically''
structured programs~\cite{gelson97a}.
A modification of the previous compilation techniques for this strategy is
discussed in \cite{delsch00b}.
Although conceptually different, one finds similar strategies when dealing
with inheritance, update and/or dynamic logic programs
\cite{bufale99a,eifisato00a,allepeprpr98a}, respectively.

While all of the aformentioned approaches remain within the same complexity
class, other approaches step up in the polynomial hierarchy
\cite{rintanen95,sakino96,zhafoo97a}.
Among them, preferences on literals are investigated in~\cite{sakino96}.
In contrast to these approaches, so-called courteous logic
programs~\cite{grosof97} step down the polynomial hierarchy into $P$.
Due to the restriction to acyclic positive logic programs a courteous answer
set can be computed in $O(n^2)$ time.
Other preference-based approaches that exclude negation as failure include
\cite{dimkak95a,pramin96,yowayu01} as well as the framework of defeasible
logics \cite{nute87,nute94}.
A comparision of the latter with preferred well-founded semantics (as defined
in~\cite{brewka96a}) can be found in \cite{brewka01a}.

In a companion paper,
we exploit our fixpoint operators for defining regular and well-founded
semantics for ordered logic programs within an alternating fixpoint theory.%
\footnote{This material was removed from this paper due to
space restrictions.}
This yields a surprising yet negative result insofar as these operators turn
out to be too weak in the setting of well-founded semantics.
We address this by defining a parameterizable framework for preferred well-founded
semantics, summarized in~\cite{schwan02b}.



\section{Conclusion}
\label{sec:conclusion}

The notion of preference seems to be pervasive in logic programming when it
comes to knowledge representation.
This is reflected by numerous approaches that aim at enhancing logic
programming with preferences in order to improve knowledge representation
capacities.
Despite the large variety of approaches, however, only very little attention
has been paid to their structural differences and sameness,
finally leading to solid semantical underpinnings.
In particular, there were up to now only few attempts to characterize one
approach in terms of another one.
The lack of this kind of investigation is clearly due to the high diversity of
existing approaches.

This work is a first step towards a systematic account to logic programming
with preferences.
To this end, we employ fixpoint operators following the tradition of logic
programming.
We elaborated upon three different approaches that were originally defined in
rather heterogenous ways.
We obtained three alternative yet uniform ways of characterizing preferred
answer sets
(in terms of fixpoints, order preservation, and an axiomatic account).
The underlying uniformity provided us with a deeper understanding of how and
which answer sets are preferred in each approach.
This has led to a clarification of their relationships and subtle differences.
On the one hand, we revealed that the investigated approaches yield an
increasing number of answer sets depending on how tight they connect
preference to groundedness.
On the other hand, we demonstrated how closely the compilation technique
developed in~\cite{descto00c} follows the specification given in the fixpoint
operation.
Also, we have shown that all considered answer sets semantics correspond to
the perfect models semantics whenever the underlying ordering stratifies the
program.

We started by formally developing a specific approach to preferred
answer sets semantics that is situated ``between'' the approaches
of~Delgrande et al.~\shortcite{descto00c} and that of Brewka and
Eiter~\shortcite{breeit99a}.
This approach can be seen as a refinement of the former approach in that it
allows to suspend preferences whenever the result of applying a preferred rule
has already been derived.
This feature avoids the overly strict prescriptive approach to preferences
pursued in~\cite{descto00c}, which may lead to the loss of answer sets.



\paragraph{Acknowledgements.}

This work was supported by the German Science Foundation (DFG) under grant
FOR~375/1-1,~TP~C.
We are grateful to the anonymous referees, although we were unable to follow
all suggestions due to severe space restrictions.



\section{Proofs}
\label{sec:proofs}

\renewcommand{\TPBiO}[4]{(\TPBo{#2}{#3}{#4})^{#1}}
\renewcommand{\TPBi}[5]{\TPBiO{#1}{#2}{#3}{#4}#5}
%
\begin{shortproof}{thm:TP:properties}
It can be directly verified from the definition of
$\TPo{\Pi}{<}{Y}$.
\quad\QED
\end{shortproof}

\begin{myproof}{thm:CP:properties}
\begin{enumerate}
\item 
$\CPXdefault\subseteq C_P(X)$:\quad
Since
$\CPXdefault=\bigcup_{i\geq 0}\TPi{i}{\Pi}{<}{X}{\emptyset}$
and $C_\Pi(X)=\Ti{i}{\reduct{\Pi}{X}}{\emptyset}$,
we need only to prove that
$\TPi{i}{\Pi}{<}{X}{\emptyset}\subseteq
      \Ti{i}{\Pi,X}{\emptyset}$ for $i\ge 0$ by using induction on $i$.
   \begin{description}
      \item[Base] 
       For $i=0$, it is obvious that 
       $\TPi{0}{\Pi}{<}{X}{\emptyset}=\emptyset \subseteq \Ti{0}{\Pi,X}{\emptyset}$.
      \item[Step]
        Assume that $\TPi{i}{\Pi}{<}{X}{\emptyset}\subseteq
        \Ti{i}{\Pi,X}{\emptyset}$,
        we want to show that
        $\TPi{i+1}{\Pi}{<}{X}{\emptyset}\subseteq \Ti{i+1}{\Pi,X}{\emptyset}$.
        In fact, if $L\in \TPi{i+1}{\Pi}{<}{X}{\emptyset}$,
        then, by Definition~\ref{def:Tp:W}, there is a rule $r$ in $\Pi$ 
        such that $L=\head{r}$, $\pbody{r}\subseteq \TPi{i}{\Pi}{<}{X}{\emptyset}$
        and $\nbody{r}\cap X=\emptyset$. 
        By induction assumption, 
        $\pbody{r}\subseteq \Ti{i}{\Pi,X}{\emptyset}$.
        Since the rule $L\LPif \pbody{r}$ is in the reduct program $P^X$,
        $L\in \Ti{i+1}{\Pi,X}{\emptyset}$.
   \end{description}
\item $ C_P(X)\subseteq \CPXdefault$ if $X\subseteq \CPXdefault$:\quad
For simplicity, we denote 
\(
T_i=\Ti{i}{\Pi,X}{\emptyset}
\)
and
\(
X_i=\TPi{\Pi}{<}{X}{\emptyset}
\)
for $i\ge 0$.
It suffices to prove $\Ti{i}{\Pi,X}{\emptyset}\subseteq \CPXdefault$ 
for $k\ge 0$ by using induction on $k$. That is, for each $i\ge 0$,
there is $n_i\ge 0$ such that
\(
T_i\subseteq X_{n_i}
\)

  \begin{description}
   \item[Base]  
     If $k=1$, it is obvious that 
     $\Ti{0}{\Pi,X}{\emptyset}=\emptyset \subseteq X_0$.
   \item[Step]
     Assume that $T_i\subseteq X_{n_i}.$
     We want to show $T_{i+1}\subseteq X_{n_{i+1}}.$
     Let $a\in T_{i+1}$, then there is a rule $r\in\Gamma$
     with $\head{r}=a$, $\pbody{r}\subseteq T_i$ and
     $\nbody{r}\cap X=\emptyset$. 
     By the induction assumption, $r$ is active wrt $(X_{n_i},X)$.
     We claim that there will be no rule $r'$ such that both of Condition I
     and II hold wrt $(X_{n_i},X)$. 
     Otherwise, suppose that there is a rule $r'$ such that 
     $\head{r'}\not\in X_{n_i}$, $r<r'$ and $r'$ is active wrt $(X,X_{n_i})$.
     Without loss of generality, there is no rule $r''$ such that
     $\head{r''}\not\in X_{n_i}$, $r<r''<r'$ and $r''$ is active wrt $(X,X_{n_i})$.
     Since $X\subseteq \CPXdefault$, there be a number $n\ge n_i$ such that
     $r'$ is active wrt $(X_n,X)$. By the assumption of $r''$,
     it should be that $\head{r''}\in X_n$. A contradiction.
     Therefore, $\head{r}\in X_{n_i+1}$.
     \end{description} 
\item
If $<$ is empty, then the condition $\mathit{II}$ in 
Definition~\ref{def:Tp:W} is automatically satisfied because,
for any rule $r\in \Pi$, there is no rule $r'$ that is preferred
to $r$. This implies that
$\TPi{i}{\Pi}{<}{X}{\emptyset} = \Ti{i}{\Pi,X}{\emptyset}$ for any $i\ge 0$.
Therefore, $\CPXdefault =  C_P(X)$.
\quad\QED
\end{enumerate}
\end{myproof}

\begin{shortproof}{thm:CP:anti:monotonicty}
If $X\subseteq X'$, it is a direct induction on $i$ to show that
\(
\TPi{i}{\Pi}{<}{X'}{\emptyset} \subseteq \TPi{i}{\Pi}{<}{X}{\emptyset}
\).%
\QED
\end{shortproof}

\begin{myproof}{thm:total:uniqueness:W}
If $\Pi$ has no consistent answer set, the conclusion is obvious. 
Thus, we assume that $X$ is consistent.
First, we can easily generalize the notion of generating rules as follows:
For any two sets $Y_1$ and $Y_2$ of literals, set
$\Gamma(Y_1,Y_2)=\{\head{r}\LPif \pbody{r} \mid 
               \pbody{r}\subseteq Y_1, \nbody{r}\cap Y_2=\emptyset\}$.

Since $X$ is an answer set of $\Pi$, we have 
$X=C_\Pi(X)=\bigcup_{i\geq 0}\Ti{i}{\Pi,X}{\emptyset}$.
Let $\Gamma_0=\Gamma(T_{\Pi^X}\emptyset, X)$ and
$\Gamma_{k+1}=\Gamma(T_{\Pi^X}^k\emptyset,X)-\Gamma_k$ for $k\ge 1$.
Define a total order $\ll_X$ on $\Pi$ such that the following requirements 
are satisfied:
   \begin{enumerate}
     \item $r'\ll_X r$ for any $r\in \Gamma_k$ and $r'\in \Gamma_{k+1}$, $k=0, 1,\ldots.$
     \item If $r\in \cup_{n\ge 0} \Gamma_n$ and $r'\not\in \cup_{n\ge 0} \Gamma_n$, then
           $r'\ll_X r$.
   \end{enumerate}
Since $\Gamma_k\cap\Gamma_{k'}\neq \emptyset$ for $n\neq n'$, 
such an ordering exists.
Denote $X_i=\TPi{i}{\Pi}{\ll_X}{X}{\emptyset}$.
We need only to prove the following two propositions
P1 and P2:
\begin{description}
\item[P1] $X$ is a prioritized answer set of $(\Pi,\ll_X)$:
Since $C_P(X)=X$, it suffices to
prove that $\CPXdefault=C_P(X)$. Firstly, by Theorem~\ref{thm:CP:properties},
$\CPXdefault\subseteq C_P(X)$. 
For the opposite inclusion, 
we note that $C_P(X)=\head{\cup_{k\ge 0} \Gamma_k)}$, where
$\head{\cup_{k\ge 0} \Gamma_k)}=\{\head{r}\;|\; r\in \cup_{k\ge 0} \Gamma_k\}$. 
Hence, we need only to prove that
$\head{\Gamma_k}\subseteq \CPXdefault$ for any $k\ge 0$ 
by using induction on $k$.
  \begin{description}
    \item[Base]
     For $k=0$, without loss of generality, suppose that
     $\Gamma_0=\{r_1,\ldots, r_t\}$ and $r_t\ll_X \cdots\ll_X r_1$. 
     We use second induction to show that $\head{r_i}\in C_P(X)$ for $1\le i\le t$.
      \begin{description}
        \item[Base] For $i=1$, since there is no rule $r'$ with $r_1\ll_X r'$, 
            $\head{r_1}\in X_1$. 
        \item[Step] Assume that $\head{r_i}\in X_i$, then 
            $\head{r_{i+1}}\in X_{i+1}$. Thus $\head{\Gamma_0}\subseteq X_t$. 
      \end{description}
    \item[Step] Assume that $\head{\Gamma_k}\subseteq \CPXdefault$. 
     Then $\head{\Gamma_k}\in X_{m_k}$ for some ${m_k}>0$. 
     Let $\Gamma_{k+1}=\{r_1,\ldots, r_u\}$ and $r_u\ll_X \cdots\ll_X r_1$.
     Then, similar to the case of $k=0$, we have that
     $\head{r_i}\in X_{{m_k}+i}$ for $i=1, \ldots, u$. 

     Thus, $\head{\Gamma_k}\subseteq \CPXdefault$ for any $k\ge 0$. 
  \end{description}
This implies that $C_P(X)\subseteq \CPXdefault$. 
Therefore, $\CPXdefault=X$.

\item [P2] If $X'$ is an answer set of $\Pi$ such that $X'\neq X$, then $X'$ is not
a prioritized answer set of $(\Pi,\ll_X)$:
First note that $X\setminus X'\neq \emptyset$ and $X'\setminus X\neq \emptyset$.
We assert that there is literal $l\in X\setminus X'$ such that $l\not\in \CPOdefault(X')$:
otherwise, $X\setminus X'\subseteq \CPOdefault(X')$. We can choose $t\ge 0$ and
a literal $l_0 \in X\setminus X'$ such that $X'_t\subseteq X\cap X'$ and $l_0\in X'_{t+1}$.
Then there is a rule $r$ such that $\head{r}=l_0$, $\pbody{r}\subseteq X'_t$ and
$\nbody{r}\cap X'=\emptyset$. This will implies that $l\in C_{\Pi^{X'}}(X')$,
i.~e. $l\in X'$, contradiction.
Therefore, we have shown that there is a rule $r$ in $\Pi$ such that $\head{r}
\in X$ and $\head{r}\not\in \CPOdefault(X')$.
For each $l'\in X'\setminus X$ and each rule $r'$ such that $\head{r'}=l'$,
we have  $r'\ll_X r$. Thus, we know that $l'\not\in \CPOdefault(X')$.
This means that $X'\neq \CPOdefault(X')$ and thus, $X'$ is not a prioritized
answer set of $(\Pi,\ll_X)$.
\quad\QED
\end{description}
\end{myproof}

\begin{myproof}{thm:total:at:most:one:W}
On the contrary, suppose that $(\Pi,<)$ has two distinct prioritized answer sets
$X$ and $X'$. 
Since $X\setminus X'\neq \emptyset$ and $X'\setminus X \neq \emptyset$,
there are literals $l$ and $l'$ such that $l\in X\setminus X'$ and 
$l'\in X'\setminus X$.
Without loss of generality, assume that 
$\TPi{i}{\Pi}{<}{X}{\emptyset}=\TPi{i}{\Pi}{<}{X'}{\emptyset}$ for 
$i\le n$ but
$l\in \TPi{n+1}{\Pi}{<}{X}{\emptyset}$ and 
$l'\in \TPi{n+1}{\Pi}{<}{X'}{\emptyset}$.
This means that there are two rules $r$ and $r'$ such that 
$\head{r}=l$, $\head{r'}=l'$,
$r$ and $r'$ satisfy the two conditions $\mathit{I}$ and $\mathit{II}$
in Definition~\ref{def:Tp:W} at stage $n$ with respect to $X$ and $X'$, 
respectively.
We observe two obvious facts:
F1. $r'$ is active wrt $(X, \TPi{n}{\Pi}{<}{X}{\emptyset})$;
and F2. $r$ is active wrt $(X', \TPi{n}{\Pi}{<}{X'}{\emptyset})$.
By F1, we have $r'\ll r$. Similarly, by F2, it should be $r\ll r'$,
contradiction. Therefore, $(\Pi,\ll)$ has the unique prioritized
answer sets.
\quad\QED
\end{myproof}

\newcommand{\pCPOdefault}{\CPO{\Pi}{<_s}}
\begin{myproof}{thm:stratified:perfect:W:i}
 \begin{enumerate}
  \item $X^\star=M_t$ is a prioritized answer set of $(\Pi,<_s)$: 
     \(
      X^\star=\pCPOdefault(X^\star)
     \).

     \begin{enumerate}
       \item $\pCPOdefault(X^\star)\subseteq X^\star$: we show that 
        $\TPi{i}{\Pi}{<_s}{X^\star}{\emptyset}\subseteq X^\star$ by using induction on $i$.
         \begin{description}
          \item[Base] 
        For $i=0$, $\TPi{0}{\Pi}{<_s}{X^\star}{\emptyset}=\emptyset\subseteq X^\star$ 
        is obvious.

          \item[Step] Assume that $\TPi{i}{\Pi}{<_s}{X^\star}{\emptyset}\subseteq X^\star$. 
        If $p\in \TPi{i+1}{\Pi}{<_s}{X^\star}{\emptyset}$, then there is a rule $r$ in
        $\Pi$ such that $p=head(r)$, 
        $\pbody{r}\subseteq \TPi{i}{\Pi}{<_s}{X^\star}{\emptyset}$ and 
        $\nbody{r}\cap X^\star=\emptyset$.
        By induction assumption, $\pbody{r}\subseteq X^\star$. 
        If $r\in \Pi_j$, then
        $\pbody{r}\subseteq M_j$ and $\nbody{r}\cap M_{j-1}=\emptyset$. 
        Therefore, $p\in X^\star$. 
        That is, $\TPi{i+1}{\Pi}{<_s}{X^\star}{\emptyset}\subseteq X^\star$.
          \end{description}
       \item $X^\star\subseteq \pCPOdefault(X^\star)$: we show that 
        $M_i\subseteq \pCPOdefault(X^\star)$ for $0\le i\le t$.
          \begin{description}
          \item[Base]  For $i=1$, it is obvious since $M_0=\emptyset$.
         \item[Step]
        If we have shown $M_i\subseteq \pCPOdefault(X^\star)$,
        we want to show that $M_{i+1}\subseteq \pCPOdefault(X^\star)$.
        We again use second induction on $k$ to prove that 
        if $p\in \Ti{k}{\Pi_{i+1},M_i}{M_i}$, then
        $p\in \pCPOdefault(X^\star)$:
        \begin{description}
          \item[Base] 
        For $k=1$, i.~e. $p\in \Ti{1}{\Pi_{i+1},M_i}{M_i}$, if $p\not\in M_i$,
        then there is a rule $r$ in $\Pi_{i+1}$ such that $p=head(r)$,
        $\pbody{r}=\emptyset$ and $\nbody{r}\cap M_i=\emptyset$. 
        Then $\nbody{r}\cap X^\star=\emptyset$.

        By the first induction assumption, 
        $M_i\subseteq \TPi{j_0}{\Pi}{<_s}{X^\star}{\emptyset}$ for some $j_0$.
        If there are $j>0$ and a rule $r'$ such that $r<_s r'$ and 
        $r'$ is active with respect to $(X^\star,\TPi{j}{\Pi}{<_s}{X^\star}{\emptyset})$ and
        $head(r')\not\in \TPi{j}{\Pi}{<_s}{X^\star}{\emptyset}$.
        Then,
        \(
        \pbody{r'}\subseteq X^\star
        \) 
        and
        \(
        \nbody{r'}\cap \TPi{j}{\Pi}{<_s}{X^\star}{\emptyset}=\emptyset
        \). 
        We assert that $j\le j_0$.
        Otherwise, if $j>j_0$, 
        $\nbody{r'}\cap \TPi{j}{\Pi}{<_s}{X^\star}{\emptyset}=\emptyset$ 
        $\Rightarrow$ $\nbody{r'}\cap \TPi{j_0}{\Pi}{<_s}{X^\star}{\emptyset}=\emptyset$ 
        $\Rightarrow$ $\nbody{r'}\cap M_i=\emptyset$
        $\Rightarrow$ $\nbody{r'}\cap X^\star=\emptyset$.
        Therefore, $head(r')\in M_i\subseteq \TPi{j}{\Pi}{<_s}{X^\star}{\emptyset}$,
        a contradiction.
        Thus, when $j>j_0$, there will be no rule in $\Pi$ that prevents $r$ 
        to be included in $\TPi{j}{\Pi}{<_s}{X^\star}{\emptyset}$.
        Thus, $p\in \pCPOdefault(X^\star)$.
            \end{description}
           \item[Step]
        Assume that $p\in \pCPOdefault(X^\star)$ if $p\in 
        \Ti{k}{\Pi_{i+1},M_i}{M_i}$. 
        Suppose that $p\in \Ti{k+1}{\Pi_{i+1},M_i}{M_i}$ but $p\not\in M_i$, 
        then there is a rule $r$ in $\Pi_{i+1}$ such that $p=head(r)$,
        $\pbody{r}\subseteq \Ti{k}{\Pi_{i+1},M_i}{\emptyset}$ 
        and $\nbody{r}\cap M_i=\emptyset$.
        Then $\pbody{r}\subseteq M_i\subseteq \TPi{j_0}{\Pi}{<_s}{X^\star}{\emptyset}$ 
        for some $j_0$ and $\nbody{r}\cap X^\star=\emptyset$.
        Similar to the proof of the case $k=1$, we can also prove that 
        $p\in \pCPOdefault(X^\star)$.
          \end{description}
     \end{enumerate}
\item If $X=\CPX{\Pi}{<_s}{X}$, then $X$ is a preferred answer set of $(\Pi,<_s)$.
By Corollary~\ref{thm:CP:Cp}, $X$ is also an answer set of $\Pi$.
However, $\Pi$ has the unique answer set $X^\star$ and thus $X=X^\star$. 
\quad\QED
\end{enumerate}
\end{myproof}

\begin{myproof}{thm:stratified:perfect:W:ii}
By Theorem~\ref{thm:stratified:perfect:W:i} (1), the perfect model
$X^\star$ is a preferred answer set. On the other hand, since each
preferred answer set $X$ is also a standard answer set. In particular,
for the stratified program $\Pi$, it has the unique answer set $X^\star$.
Therefore, $X=X^\star$.
\quad\QED
\end{myproof}

\begin{myproof}{thm:CP:C:W}
Let $(\Pi,<)$ be an ordered logic program over $\mathcal{L}$
and
$X$ a consistent set of literals over $\mathcal{L}$.
\paragraph{``$\subseteq$''-part}
Define\footnote{As defined in Section~\ref{sec:delgrande:schaub:tompits},
  \rul{\cdot} is a bijective mapping between rule heads and rules.}
\begin{eqnarray*}
  Y &=&\quad \{\head{r} \mid r\in\rul{C_{\Trans{\Pi,<}}(Y)}\}
  \\& &\cup\;\{\applied{\name{r}}\mid r    \in\rul{C_{\Trans{\Pi,<}}(Y)}\}
       \cup  \{\blocked{\name{r}}\mid r\not\in\rul{C_{\Trans{\Pi,<}}(Y)}\}
  \\& &\cup\;\{\ok{\name{r}}   \mid r\in\Pi\}
       \cup  \{\oko{\name{r}}{\name{r'}}\mid r,r'\in\Pi\}
\end{eqnarray*}
Clearly, we have $X=Y\cap\mathcal{L}$.
By definition, we have
\(
\CPXdefault=\bigcup_{i\geq 0}\TPi{i}{\Pi}{<}{X}{\emptyset}
\)
and
\(
C_{\Trans{\Pi,<}}(Y) = \Cn{\reduct{\Trans{\Pi,<}}{Y}}
\).

In view of this,
we show by induction that
\(
\TPi{i}{\Pi}{<}{X}{\emptyset}\subseteq \Cn{\reduct{\Trans{\Pi,<}}{Y}}
\)
for ${i\geq 0}$.
To be precise, we show for every $r\in\Pi$ by nested induction that
\(
\head{r}\in\TPi{i}{\Pi}{<}{X}{\emptyset}
\)
implies
\(
\head{r}\in\Cn{\reduct{\Trans{\Pi,<}}{Y}}
\)
and moreover, for every $r'\in\Pi$,
that
if
\(
r<r'
\)
then
\(
\blocked{\name{r'}}\in\Cn{\reduct{\Trans{\Pi,<}}{Y}}
\)
or
\(
\applied{\name{r'}}\in\Cn{\reduct{\Trans{\Pi,<}}{Y}}
\)
or
\(
\head{\name{r'}}\in\Cn{\reduct{\Trans{\Pi,<}}{Y}}
\).
\paragraph{$i=0$}
By definition,
\(
\TPi{0}{\Pi}{<}{X}{\emptyset}=\emptyset\subseteq \Cn{\reduct{\Trans{\Pi,<}}{Y}}
\).
\paragraph{$i>0$}
Consider $r\in\Pi$ such that
\(
\head{r}\in\TPi{i+1}{\Pi}{<}{X}{\emptyset}
\).
By definition, we have that $r$ is active wrt
\(
(\TPi{i}{\Pi}{<}{X}{\emptyset},X)
\).
That is,
\begin{enumerate}
\item\label{lab:forty:six} $\pbody{r}\subseteq\TPi{i}{\Pi}{<}{X}{\emptyset}$.
  By the induction hypothesis, we get
  \(
  \pbody{r}\subseteq \Cn{\reduct{\Trans{\Pi,<}}{Y}}
  \).
\item\label{lab:forty:svn} $\nbody{r}\cap X=\emptyset$.
  By definition of $Y$, this implies
  \(
  \nbody{r}\cap Y=\emptyset
  \).

  Furthermore, this implies that
  \(
  \reductr{\ap{2}{r}}
  =
  \applied{\name{r}}\LPif{}\ok{\name{r}},\pbody{r}
  \in
  \reduct{\Trans{\Pi,<}}{Y}
  \).
\end{enumerate}
We proceed by induction on $<$.
\begin{description}
\item[Base] Suppose $r$ is maximal with respect to $<$.
  We can show the following lemma.
  \begin{lemma}\label{lem:CP:C:W:two}
    If $r\in\Pi$ is maximal with respect to $<$,
    then
    \(
    \ok{\name{r}}\in\Cn{\reduct{\Trans{\Pi,<}}{Y}}
    \).
  \end{lemma}
  Given that we have just shown in~\ref{lab:forty:six} and~\ref{lab:forty:svn}
  that
  \(
  \pbody{r}\subseteq\Cn{\reduct{\Trans{\Pi,<}}{Y}}
  \)
  and
  \(
  \reductr{\ap{2}{r}}
  \in
  \reduct{\Trans{\Pi,<}}{Y}
  \),
  Lemma~\ref{lem:CP:C:W:two} and the fact that
  \Cn{\reduct{\Trans{\Pi,<}}{Y}} is closed under
     {\reduct{\Trans{\Pi,<}}{Y}}
  imply that
  \(
  \applied{\name{r}}\in\Cn{\reduct{\Trans{\Pi,<}}{Y}}.
  \)
  Analogously, we get
  \(
  \head{r}\in\Cn{\reduct{\Trans{\Pi,<}}{Y}}
  \)
  due to 
  \(
  \reductr{\ap{1}{r}}
  \in
  \reduct{\Trans{\Pi,<}}{Y}
  \).
  We have thus shown that
  \(
  \{\head{r},\applied{\name{r}}\}
  \subseteq
  \Cn{\reduct{\Trans{\Pi,<}}{Y}}
  \).
\item[Step] 
  We start by showing the following auxiliary result.
  \begin{lemma}\label{lem:CP:C:W:one}
    Given the induction hypothesis,
    we have
    \(
    \ok{\name{r'}}\in\Cn{\reduct{\Trans{\Pi,<}}{Y}}
    \). 
  \end{lemma}
  \begin{interproof}{lem:CP:C:W:one}
    Consider $r''\in\Pi$ such that $r'<r''$.
    By the induction hypothesis,
    we have either
    \(
    \blocked{\name{r''}}\in\Cn{\reduct{\Trans{\Pi,<}}{Y}}
    \)
    or
    \(
    \applied{\name{r''}}\in\Cn{\reduct{\Trans{\Pi,<}}{Y}}
    \)
    or
    \(
    \head{\name{r''}}\in\Cn{\reduct{\Trans{\Pi,<}}{Y}}
    \).
    Clearly,
    we have
    \(
    (\PREC{\name{r'}}{\name{r''}})\in
    \Cn{\reduct{\Trans{\Pi,<}}{Y}}
    \)
    iff $r'<r''$.
    Hence, whenever $r'<r''$, we obtain
    \(
    \oko{\name{r'}}{\name{r''}}
    \in
    \Cn{\reduct{\Trans{\Pi,<}}{Y}}
    \)
    by means of \reductr{\cok{3}{r'}{r''}}, \reductr{\cok{4}{r'}{r''}}, or
    \reductr{\cok{5}{r'}{r''}}
    (all of which belong to ${\reduct{\Trans{\Pi,<}}{Y}}$).
    Similarly, we get 
    \(
    \oko{\name{r'}}{\name{r''}}
    \in
    \Cn{\reduct{\Trans{\Pi,<}}{Y}}
    \),
    whenever $r'\not<r''$ from \reductr{\cok{2}{r'}{r''}}.
    Lastly, we obtain
    \(
    \ok{\name{r'}}
    \in
    \Cn{\reduct{\Trans{\Pi,<}}{Y}}
    \)
    via $\reductr{\cokt{1}{r'}}\in{\reduct{\Trans{\Pi,<}}{Y}}$.
    \quad\QED
  \end{interproof}
  For all rules $r'$ with $r<r'$, we have that either
  \begin{enumerate}
  \item $r'$ is not active wrt
    \(
    (X,\TPi{i}{\Pi}{<}{X}{\emptyset})
    \).
    That is, we have that either
    \begin{enumerate}
    \item\label{lab:forty:two} $\pbody{r}\not\subseteq X$.
      By definition of $Y$, this implies
      \(
      \pbody{r}\not\subseteq Y
      \).

      By definition,
      \(
      \reductr{\bl{1}{r'}{L^{+}}}
      =
      \blocked{\name{r'}}\LPif\ok{\name{r'}}
      \in
      \reduct{\Trans{\Pi,<}}{Y}
      \)
      for some $L^{+}\in\pbody{r}$ such that $L^{+}\not\in Y$.
      By Lemma~\ref{lem:CP:C:W:one},
      we have
      \(
      \ok{\name{r'}}\in\Cn{\reduct{\Trans{\Pi,<}}{Y}}
      \).
      Given that \Cn{\reduct{\Trans{\Pi,<}}{Y}} is closed under
      {\reduct{\Trans{\Pi,<}}{Y}},
      we get that
      \(
      \blocked{\name{r'}}\in\Cn{\reduct{\Trans{\Pi,<}}{Y}}
      \).
    \item $\nbody{r}\cap\TPi{i}{\Pi}{<}{X}{\emptyset}\neq\emptyset$.
      By the induction hypothesis, this implies that
      \(
      \nbody{r}\cap \Cn{\reduct{\Trans{\Pi,<}}{Y}}\neq\emptyset
      \).

      Therefore,
      \(
      \reductr{\bl{2}{r}{L^{-}}}
      =
      \blocked{\name{r}}\LPif\ok{\name{r}},L^{-}
      \in
      \reduct{\Trans{\Pi,<}}{Y}      
      \)
      for some $L^{-}\in\nbody{r}\cap \Cn{\reduct{\Trans{\Pi,<}}{Y}}$.
      In analogy to~\ref{lab:forty:two}, this allows us to conclude that
      \(
      \blocked{\name{r'}}\in\Cn{\reduct{\Trans{\Pi,<}}{Y}}
      \).
    \end{enumerate}
    In both cases, we conclude
    \(
    \blocked{\name{r'}}\in\Cn{\reduct{\Trans{\Pi,<}}{Y}}
    \).
    By the induction assumption, 
    \(
    {\head{r'}}\in\Cn{\reduct{\Trans{\Pi,<}}{Y}}
    \).
  \end{enumerate}
  We have thus shown that either
  \(
  \blocked{\name{r'}}\in\Cn{\reduct{\Trans{\Pi,<}}{Y}}
  \)
  or
  \(
  {\head{r'}}\in\Cn{\reduct{\Trans{\Pi,<}}{Y}}
  \)
  for all $r'$ such that $r<r'$.

  In analogy to what we have shown in the proof of Lemma~\ref{lem:CP:C:W:one},
  we can now show that
  \(
  \ok{\name{r}}\in\Cn{\reduct{\Trans{\Pi,<}}{Y}}
  \).

  In analogy to the base case, we may then conclude
  \(
  \{\head{r},\applied{\name{r}}\}
  \subseteq
  \Cn{\reduct{\Trans{\Pi,<}}{Y}}
  \).
\end{description}
\paragraph{``$\supseteq$''-part}
We have $X=Y\cap\mathcal{L}$.
By definition, we have
\(
C_{\Trans{\Pi,<}}(Y) = \Cn{\reduct{\Trans{\Pi,<}}{Y}}
\)
and moreover that
\(
\Cn{\reduct{\Trans{\Pi,<}}{Y}}
=
\bigcup_{i\geq 0}\Ti{i}{\reduct{\Trans{\Pi,<}}{Y}}{\emptyset}
\).
Given this, 
we show by induction that
\(
(\Ti{i}{\reduct{\Trans{\Pi,<}}{Y}}{\emptyset}\cap\mathcal{L})
\subseteq
\CPXdefault
\)
for ${i\geq 0}$.
\paragraph{$i=0$}
By definition,
\(
\Ti{0}{\reduct{\Trans{\Pi,<}}{Y}}{\emptyset}
=\emptyset
\subseteq\CPXdefault
\).
\paragraph{$i>0$}
Consider $r\in\Pi$ such that
\(
\head{r}\in(\Ti{i+1}{\reduct{\Trans{\Pi,<}}{Y}}{\emptyset}\cap\mathcal{L})
\).
In view of {\reduct{\Trans{\Pi,<}}{Y}},
this implies that
\(
\applied{\name{r}}\in(\Ti{i}{\reduct{\Trans{\Pi,<}}{Y}}{\emptyset}\cap\mathcal{L})
\)
and thus
\(
\reductr{\ap{2}{r}}\in{\reduct{\Trans{\Pi,<}}{Y}}
\).
The latter implies that
\(
\nbody{r}\cap Y=\emptyset
\),
whence
\(
\nbody{r}\cap X=\emptyset
\)
because of $X=Y\cap\mathcal{L}$.
The former implies that
\(
\pbody{r}\cup\{\ok{\name{r}}\}
\subseteq
\Ti{i-1}{\reduct{\Trans{\Pi,<}}{Y}}{\emptyset}
\).
By the induction hypothesis, we obtain that
\(
\pbody{r}\subseteq\CPXdefault
\).
Consequently, $r$ is active wrt
\(
(\CPXdefault,X)
\).

Suppose there is some $r'\in\Pi$ with $r<r'$ such that
\begin{enumerate}
\item $r'$ is active wrt $(X,\CPXdefault)$.
That is,
\begin{enumerate}
\item \label{lab:forty:for}$\pbody{r}\subseteq X$ and
\item \label{lab:forty:fif}$\nbody{r}\cap\CPXdefault=\emptyset$.
\end{enumerate}
\item\label{lab:forty:tri} $\head{r'}\not\in\CPXdefault$.
\end{enumerate}
By the induction hypothesis, we obtain from~\ref{lab:forty:tri} that
\(
\head{r'}\not\in\Ti{j}{\reduct{\Trans{\Pi,<}}{Y}}{\emptyset}
\)
for ${j\leq i}$.

Clearly, 
we have
\(
(\PREC{\name{r'}}{\name{r''}})\in\Ti{i}{\reduct{\Trans{\Pi,<}}{Y}}{\emptyset}
\)
for $i\geq 1$
iff $r'<r''$.
Moreover,
\(
\ok{\name{r}}
\in
\Ti{i-1}{\reduct{\Trans{\Pi,<}}{Y}}{\emptyset}
\)
implies (see above)
\(
\oko{\name{r}}{\name{r''}}
\in
\Ti{i-2}{\reduct{\Trans{\Pi,<}}{Y}}{\emptyset}
\)
for all $r''\in\Pi$.
This and the fact that
\(
\head{r'}\not\in\Ti{j}{\reduct{\Trans{\Pi,<}}{Y}}{\emptyset}
\)
for ${j\leq i}$ implies that
\(
\blocked{\name{r'}}
\in
\Ti{i-3}{\reduct{\Trans{\Pi,<}}{Y}}{\emptyset}
\).

This makes us distinguish the following two cases.
\begin{enumerate}
\item If $\blocked{\name{r'}}$ is provided by ${\bl{1}{r'}{L^{+}}}$,
  then there is some $L^{+}\in\pbody{r'}$ such that
  \(
  L^{+}\not\in Y
  \).
  Given that $X=Y\cap\mathcal{L}$, this contradicts~\ref{lab:forty:for}.
\item If $\blocked{\name{r'}}$ is provided by ${\bl{2}{r'}{L^{-}}}$,
  then there is some $L^{-}\in\nbody{r'}$ such that
  \(
  L^{-}\in\Ti{i-4}{\reduct{\Trans{\Pi,<}}{Y}}{\emptyset}
  \).
  By the induction hypothesis, we obtain that
  \(
  L^{-}\in\CPXdefault
  \).
  A contradiction to~\ref{lab:forty:fif}.
\end{enumerate}
So, given that
$r$ is active wrt
\(
(\CPXdefault,X)
\)
and that there is no $r'\in\Pi$ such that $r<r'$
satisfying~\ref{lab:forty:for}, \ref{lab:forty:fif}, and \ref{lab:forty:tri},
we have that
\(
\head{r}
\in
\TP{\Pi}{<}{X}{(\CPXdefault)}
\).
That is,
\(
\head{r}
\in
\CPXdefault
\).
\quad\QED
\end{myproof}

\begin{shortproof}{thm:CP:T:W}
It follows from Lemma~\ref{lem:preserving:W} and Lemma~\ref{lem:results:W:2e3}.
\quad\QED
\end{shortproof}

%
\begin{shortproof}{thm:stratified:perfect:D}
Similar to the proof of Theorem~\ref{thm:stratified:perfect:W:i}.
\quad\QED
\end{shortproof}

By Theorem~4.8 in~\cite{descto02a}, it suffices to show the following
Lemma~\ref{lem:preserving:D}. Before doing this, we first present a definition.
Given a statically ordered logic program $(\Pi,<)$ and
a set $X$ of literals, set
\renewcommand{\TPo}[3]{\ensuremath{(\mathcal{T}^{\DST})_{(#1,#2),#3}}}
\renewcommand{\TPDiO}[4]{\TPo{#2}{#3}{#4}^{#1}}
\(
X_i=\TPDi{i}{\Pi}{<}{X}{\emptyset}
\)
for $i\ge 0$.
%
\begin{definition}
Let $(\Pi,<)$ be a statically ordered logic program and $r$ be a rule in $\Pi$.
$X$ and $X_i (i\ge 1)$ are as above.
We say another rule $r'$ is a \emph{\DST-preventer} of $r$ in the context $(X,X_i)$
if 
(1) $r< r'$ and
(2) $r'$ is active wrt $(X,X_i)$ and $r'\not\in \rul{X_i}$.
\end{definition}
%
%
\begin{lemma} \label{lem:pre:D}
Let $(\Pi,<)$ be a statically ordered logic program and $X$ a set of literals
with 
    \(
    \mathcal{C}^{\DST}_{(\Pi,<)}(X)= X
    \).
Then, for any $r\in \GR{\Pi}{X}$, there exists a number $i$ such that
$r\in \rul{X_i}$.
\end{lemma}
The intuition behind this lemma is that each \DST-preventer of a rule in
$\GR{\Pi}{X}$ is a ``temporary'' one if
    \(
    \mathcal{C}^{\DST}_{(\Pi,<)}(X)= X
    \).
\begin{interproof}{lem:pre:D}
On the contrary, suppose that there is a rule $r\in \GR{\Pi}{X}$ such that
$r\not\in \rul{X_i}$ for any $i$. Without loss of generality, 
assume that there is no such rule that is preferred than $r$. 

Since $r\in \GR{\Pi}{X}$ and $X=\cup_{i=1}^{\infty}X_i$, $r$ will become 
active wrt $(X_t, X)$ at some stage $t\ge 0$. Therefore, it must be the
case that there is a \DST-preventer $r'$ satisfying $r'\in \GR{\Pi}{X}$.
This implies that $r<r'$ and $r'\in \GR{\Pi}{X}$ but 
$r'\not\in \rul{X_i}$ for any $i$, contradiction to our assumption on $r$.
Thus, the lemma is proven.
\quad\QED
\end{interproof}
%
\begin{lemma} \label{lem:preserving:D}
  Let $(\Pi,<)$ be a statically ordered logic program and $X$ a set of literals.
Then  
  $X$ is a $<^\DST$-preserving answer set of\/$\Pi$ if and only if $X$ is a
  set of literals with
    \(
    \mathcal{C}^{\DST}_{(\Pi,<)}(X)= X
    \).
\end{lemma}
\begin{interproof}{lem:preserving:D}
Without loss of generality, assume that 
$\rul{X_i}=\{r_{i1},\ldots,r_{in_i}\}$ for
$i\ge 1$.
\paragraph{if part} 
Let \(
    \mathcal{C}^{\DST}_{(\Pi,<)}(X)= X
    \).
By Lemma \ref{lem:pre:D}, $\GR{\Pi}{X}=\cup_{i=1}^{\infty} \rul{X_i}$.
This means that the sequence $\Delta$:
\(
\langle r_{11},\ldots,r_{1n_1},r_{21},\ldots,r_{2n_2},\ldots\rangle
\)
is an enumeration of $\GR{\Pi}{X}$.

It suffices to prove that this sequence of rules in $\Delta$
is $<^\DST$-preserving with respect to $X$.

We need to justify the two conditions of $<^\DST$-sequence are satisfied 
by $\Delta$:
\begin{description}
\item[C1]
For each $r_i\in \rul{X_t}$ where $t>0$, then $r_i$ is active wrt
$(X_{t-1},X)$. This implies that $\pbody{r_i}\subseteq \{\head{r_j}
\mid j<i\}$. 
\item[C2] 
if $r<r'$, then $r'$ is prior to $r$ in $\Delta$: 
notice that, since $X=\cup_{i=1}^{\infty}X_i$, if a rule is active
wrt $(X_i,X)$ then it is also active wrt $(X,X_i)$.
Thus, by Definition \ref{def:Tp:W},
$r$ and $r'$ can not be in the same section $\rul{X_s}$. 
If C2 is not satisfied by $\Delta$, then there are two rules,
say $r$ and $r'$, such that $r<r'$ but $r$ is prior to $r'$ in $\Delta$.
Without loss of generality, assume that $r\in \rul{X_i}$ and $r'\in \rul{X_j}$
but $i<j$. Then $r'$ should prevent $r$ to be included in $\rul{X_i}$,
which means $r\not\in \rul{X_i}$, contradiction. Therefore, C2 holds.
\item[C3] 
if $r_i<r'$ and $r'\in \Pi\setminus \GR{\Pi}{X}$, then
$\pbody{r'}\not\subseteq X$ or $r'$ is defeated by the set $\{\head{r_j}
\mid j<i\}$:
Assume that $r_i\in \rul{X_t}$, then $r'\not\in \rul{X_t}$.
On the contrary, assume that 
$\pbody{r'}\subseteq X$ and  $r'$ is not defeated by the set
$\{\head{r_j}\mid j<i\}$, then $r'$ is not defeated by
$X_{t-1}$ because $X_{t-1}\subseteq \{\head{r_j}\mid j<i\}$.
Thus, $r'$ is active wrt $(X,X_{t-1})$ and $r'\not\in \rul{X_{t-1}}$. 
This means that $r'$ is a \DST-preventer of $r_i$ in the context
$(X,X_{t-1})$ and thus, $r_i\not\in \rul{X_t}$, contradiction.
That is, $\pbody{r'}\not\subseteq X$ or $r'$ is defeated by the set
$\{\head{r_j}\mid j<i\}$. 
\end{description}

\paragraph{only-if part} 
Assume that $X$ is a $<^\DST$-preserving
answer set of $\Pi$, then there is a grounded
enumeration $\langle r_i\rangle_{i\in I}$ of $\GR{\Pi}{X}$ such that, 
for every $i,j\in I$, we have that:
  \begin{enumerate}
    \item if $r_i<r_j$, then $j<i$;
          \ and
    \item if
          $\PRECM{r_i}{r'}$
          and 
          \(
          r'\in {\Pi\setminus\GR{\Pi}{X}},
          \)
          then
          either (a) $\pbody{r'}\not\subseteq X$
            or
           (b) $\nbody{r'}\cap\{\head{r_j}\mid j<i\}\neq\emptyset$.
        \end{enumerate}

A set $\bar{\Delta}$ of rules is {\em discrete} if there is no pair of 
rules $r$ and $r'$ in $\bar{\Delta}$ s.~t. $r<r'$.

We define recursively a sequence of sets of rules as follows.

Define $\bar{\Delta}_1$ as the largest section of 
$\langle r_i\rangle_{i\in I}$ satisfying the following conditions:
\begin{enumerate}
\item  $\bar{\Delta}_1$ is discrete;
\item  $r_1\in \bar{\Delta}_1$;
\item  $\body{r}=\emptyset$ for any $r\in \bar{\Delta}_1$.
\end{enumerate}

Suppose that $\bar{\Delta}_i$ is well-defined and $r_{m_i}$ is the last rule
of $\bar{\Delta}_i$, we define $\bar{\Delta}_{i+1}$
as the largest section of $\langle r_i\rangle_{i\in I}$ satisfying
the following conditions:
\begin{enumerate}
\item  $\bar{\Delta}_{i+1}$ is discrete;
\item  $r_{m_i+1}\in \bar{\Delta}_{i+1}$;
\item  $\body{r}\subseteq \{head(r')\mid r'\in \cup_{j=0}^i \bar{\Delta}_j\}$ 
for any $r\in \bar{\Delta}_{i+1}$.
\item  disjoint with $\cup_{j=0}^i \bar{\Delta}_j$.
\end{enumerate}

Denote $\bar{X}_i= \{head(r)\mid r\in \cup_{j=0}^i \bar{\Delta}_j\}$. Then
we have the following fact: 

{\em if $r\in \bar{\Delta}_{i+1}$ such that $\bar{X}_{i-1}\models \pbody{r}$
and no rule $r'\in \bar{\Delta}_i$ with $r<r'$, then we can move $r$ from
$\bar{\Delta}_{i+1}$ to $\bar{\Delta}_i$, the resulting sequence of rules
still is $<^\DST$-preserving.
}

Without loss of generality, assume that our sequence 
$\langle\bar{\Delta}_i\rangle$ is fully
transformed by the above transformation.
Since $\cup_{i=0}^{\infty} \bar{X}_i=X$, we can prove 
$\cup_{i=0}^{\infty} X_i=X$ by proving $\bar{X}_i=X_i$ for every 
$i\in I$.
Thus, it suffices to prove $\bar{\Delta}_i=\rul{X_i}$
for every $i\in I$. 
We use induction on $i$:
\begin{description}
\item[Base]  $\bar{\Delta}_0=\rul{X_0} =\emptyset$.
\item[Step]  Assume that $\bar{\Delta}_i=\rul{X_i}$,
we want to prove $\bar{\Delta}_{i+1}=\rul{X_{i+1}}$.

$\bar{\Delta}_{i+1}\subseteq \rul{X_{i+1}}$: 
For any $r_t\in \bar{\Delta}_{i+1}$, by the condition 3 in the 
construction of $\bar{\Delta}_{i+1}$, $r_t$ 
is active wrt $(X_i,X)$. And for any $r'$ such that $r_t<r'$
and $r'$ is active wrt $(X,X_i)$ then $\pbody{r'}\subseteq X$
and $r'$ is not defeated by $X_i$. By induction, 
$\cup_{k<t}head(r_k)\subseteq X_i=\bar{X}_i$, thus $r'$ is not defeated by
$\cup_{k<t}head(r_k)$. By Definition~\ref{def:order:preservation:D}, 
it should be the case that $r'\in\GR{\Pi}{X}$, which implies that 
$r'\in\bar{\Delta}_i=\rul{X_i}$. Therefore, $r'$ is not a \DST-preventer
of $r_t$. That is, $r_t\in \rul{X_{i+1}}$. 

$\rul{X_{i+1}}\subseteq \bar{\Delta}_{i+1}$: 
For $r\in \rul{X_{i+1}}$, we claim that $r\in  \bar{\Delta}_{i+1}$.
Otherwise, there would exist $t>i+1$ such that $r\in \bar{\Delta}_t$.
Notice that, by induction assumption, $\pbody{r}\subseteq X_i$.
Thus, it must be the case that there is at least one rule 
$r'\in \cup_{j=i+1}^{t-1} \rul{X_j}$ such that $r'<r$. But
$r'$ is active wrt $(X,X_{i+1})$, which contradicts to 
$r\in \rul{X_{i+1}}$. Therefore, $\rul{X_{i+1}}\subseteq \bar{\Delta}_{i+1}$. 
\quad\QED
\end{description}
\end{interproof}

\begin{shortproof}{thm:CP:C:D}
Similar to the proof of Theorem~\ref{thm:CP:C:W}.
\quad\QED
\end{shortproof}

%
\newcommand{\redwy}{\reduct{\Trans{\Pi,<}}{Y}}
\begin{shortproof}{thm:results:W}
It follows from the following Lemma~\ref{lem:preserving:W} and
Theorem~\ref{thm:CP:T:W}.
\quad\QED
\end{shortproof}
%
\begin{lemma}\label{lem:compilation:basic}
 Let $(\Pi,<)$ be an ordered logic program over $\mathcal{L}$
  and let $Y$ be a consistent answer set of $\TPW(\Pi,<)$. 
  Denote $X=Y\cap \mathcal{L}$.  
  Then, we have for any $r\in\Pi$:
  \begin{enumerate}
    \item $\ok{n_r}\in Y$; and
    \item $\applied{n_r}\in Y$ iff $\blocked{n_r}\not\in Y$.
  \end{enumerate}
\end{lemma}
\begin{interproof}{lem:compilation:basic}
We prove the two propositions by parallel induction on ordering
$<$.
\paragraph{Base} Let $r$ be a maximal element of $<$.
      \begin{enumerate}
        \item By assumption, $r\not<r'$ for any $r'\in\Pi$. This implies that
              $\oko{n_r}{n_{r'}}\in Y$ for any $r'\in\Pi$.
              Thus, $\ok{n_r}\in Y$.
        \item There are two possible cases:
             \begin{itemize}
                \item $\body{r}$ is satisfied by $X$: 
                   Since $\ap{2}{r}\in\redwy$, we have $\applied{n_r}\in Y$.
                \item $\body{r}$ is not satisfied by $X$:
                   The body of at least one of $\bl{1}{r}{L^{+}}$ and $\bl{2}{r}{L^{-}}$
                   is satisfied by $Y$, thus $\blocked{n_r}\in\redwy$.
             \end{itemize}
      \end{enumerate}
\paragraph{Step}
       \begin{enumerate}
        \item Consider $r\in\Pi$. Assume that $\ok{n_{r'}}\in Y$ and
       either $\applied{n_{r'}}\in Y$ or $\blocked{n_{r'}}\in Y$ for all $r'$ with $r<r'$.
       In analogy to the base case, we have $\oko{n_r}{n_{r'}}\in Y$ 
       for all $r'\in\Pi$ with $r\not<r'$. 
       
       For $r'$ with $r<r'$, by the induction assumption, we have 
        either $\applied{n_{r'}}\in Y$ or $\blocked{n_{r'}}\in Y$. Hence the body of at
        least one of $\cok{3}{r}{r'}$ and $\cok{4}{r}{r'}$ is satisfied by $Y$. 
        This implies $\oko{n_r}{n_{r'}}\in Y$.

        So, we have proved that $\oko{n_r}{n_{r'}}\in Y$ for any $r'\in\Pi$.
        Thus, $\ok{n_r}\in Y$.
        \item Analogous to the base case.
          \quad\QED
        \end{enumerate}  
\end{interproof}
          ==================================================
Given a statically ordered logic program $(\Pi,<)$ and
a set $X$ of literals, set
\renewcommand{\TPo}[3]{\ensuremath{(\mathcal{T}^{\WZL})_{(#1,#2),#3}}}
\renewcommand{\TPWiO}[4]{\TPo{#2}{#3}{#4}^{#1}}
%
\(
X_i=\TPWi{i}{\Pi}{<}{X}{\emptyset}
\)
for $i\ge 0$ and
\(
\ugr{X_i}=\{r\in\GR{\Pi}{X}\setminus\ugr{X_{i-1}}\mid\text{either $r$ applied in producing 
$X_i\setminus X_{i-1}$ or }
\head{r}\in X_{i-1}\}
\) for $i>0$. 
Intuitively, $\ugr{X_i}$ is the set of the generating rules that are used
at stage $i$.
%
\begin{definition}
Let $(\Pi,<)$ be a statically ordered logic program and $r$ be a rule in $\Pi$.
$X$ and $X_i (i\ge 0)$ are as above.
We say another rule $r'$ is a \emph{\WZL-preventer} of $r$ in the context $(X,X_i)$
if the following conditions are satisfied:
\begin{enumerate}
\item $r< r'$ and
\item $r'$ is active wrt $(X,X_i)$ and $\head{r'}\not\in X_i$.
\end{enumerate}
\end{definition}
\begin{lemma} \label{lem:pre:W}
Let $(\Pi,<)$ be a statically ordered logic program and $X$ a set of literals
with 
    \(
    \mathcal{C}^{\WZL}_{(\Pi,<)}(X)= X
    \).
Then, for any $r\in \GR{\Pi}{X}$, there exists a number $i$ such that
$r\in \ugr{X_i}$.
\end{lemma}
The intuition behind this lemma is that each \WZL-preventer of a rule in
$\GR{\Pi}{X}$ is a ``temporary'' one if
    \(
    \mathcal{C}^{\WZL}_{(\Pi,<)}(X)= X
    \).
\begin{interproof}{lem:pre:W}
On the contrary, suppose that there is a rule $r\in \GR{\Pi}{X}$ such that
$r\not\in \ugr{X_i}$ for any $i$. Without loss of generality, 
assume that there is no such rule that is preferred than $r$. 
Since $r\in \GR{\Pi}{X}$ and $X=\cup_{i=1}^{\infty}X_i$, $r$ will become 
active wrt $(X_t, X)$ at some stage $t\ge 0$. Therefore, it must be the
case that there is a \WZL-preventer $r'$ satisfying $r'\in \GR{\Pi}{X}$.
This implies that $r<r'$ and $r'\in \GR{\Pi}{X}$ but 
$r'\not\in\ugr{X_i}$ for any $i$, contradiction to our assumption on $r$.
Thus, the lemma is proven.
\quad\QED
\end{interproof}
%
\begin{lemma} \label{lem:preserving:W}
  Let $(\Pi,<)$ be a statically ordered logic program and $X$ a set of literals.
Then  
  $X$ is a $<^\WZL$-preserving answer set of\/ $\Pi$ if and only if $X$ is a
  set of literals with
    \(
    \mathcal{C}^{\WZL}_{(\Pi,<)}(X)= X
    \).
\end{lemma}
%
\begin{interproof}{lem:preserving:W}
Without loss of generality, assume that 
$\ugr{X_i}=\{r_{i1},\ldots,r_{in_i}\}$ for
$i\geq 1$.

\paragraph{if part} 
Let \(
    \mathcal{C}^{\WZL}_{(\Pi,<)}(X)= X
    \).
By Lemma \ref{lem:pre:W}, $\GR{\Pi}{X}=\cup_{i=1}^{\infty} \ugr{X_i}$.
This means that the sequence $\Delta$:
\(
\langle r_{11},\ldots,r_{1n_1},r_{21},\ldots,r_{2n_2},\ldots\rangle
\)
is an enumeration of $\GR{\Pi}{X}$.
It suffices to prove that this sequence
is $<^\WZL$-preserving with respect to $X$.

We need to justify that the three conditions of $<^\WZL$-sequence are satisfied 
by $\Delta$:
\begin{description}
\item[C1]
For each $r_i\in\Delta$, either $r_i$ is active wrt $(X_t,X)$ or $\head{r_i}\in X_t$
for some $t>0$. Thus, Condition 1 in Definition~\ref{def:order:preservation:W}
is satisfied.

\item[C2] 
If $r<r'$, then $r'$ is prior to $r$ in $\Delta$: 
Notice that $X=\cup_{i=1}^{\infty}X_i$, if a rule is active
wrt $(X_i,X)$ then it is also active wrt $(X,X_i)$.
Thus, by Definition~\ref{def:Tp:W},
$r$ and $r'$ can not be in the same section $\ugr{X_s}$. 

If C2 is not satisfied by $\Delta$, then there are two rules,
say $r$ and $r'$, such that $r<r'$ but $r$ is prior to $r'$ in $\Delta$.
Without loss of generality, assume that $r\in \ugr{X_i}$ and $r'\in \ugr{X_j}$
but $i<j$. Then $r'$ should prevent $r$ to be included in $\ugr{X_i}$,
which means $r\not\in \ugr{X_i}$, contradiction. Therefore, C2 holds.

\item[C3] 
On the contrary, suppose that Condition 3 in Definition~\ref{def:order:preservation:W}
is not satisfied. That is, there are two rules $r_i$ and $r'$ such that $r_i<r'$,
$r'\in \Pi\setminus \GR{\Pi}{X}$ and the following items hold:
  \begin{enumerate}
   \item $\pbody{r'}\subseteq X$,
   \item $r'$ is not defeated by the set $\{\head{r_j}\mid j<i\}$,
   \item $\head{r'}\not\in\{\head{r_j}\mid j<i\}$.
  \end{enumerate}
Without loss of generality, assume that $r_i\in\ugr{X_t}$, then $r'$ is not defeated by
$X_{t-1}$ because $X_{t-1}\subseteq \{\head{r_j}\mid j<i\}$.
Thus, $r'$ is active wrt $(X,X_{t-1})$ and $r'\not\in \ugr{X_{t-1}}$. 
This means that $r'$ is a \WZL-preventer of $r_i$ in the context
$(X,X_{t-1})$ and thus, $r_i\not\in \ugr{X_t}$, contradiction.
\end{description}

\paragraph{only-if part} 
Assume that $X$ is a $<^\WZL$-preserving answer set of\/ $\Pi$, then there is a grounded
enumeration $\langle r_i\rangle_{i\in I}$ of $\GR{\Pi}{X}$ such that
the three conditions in Definition~\ref{def:order:preservation:W} are
all satisfied.

A set $\bar{\Delta}$ of rules is {\em discrete} if there is no pair of 
rules $r$ and $r'$ in $\bar{\Delta}$ such that $r<r'$.
We define recursively a sequence of sets of rules as follows.

Define $\bar{\Delta}_1$ as the largest section of 
$\langle r_i\rangle_{i\in I}$ satisfying the following conditions:
\begin{enumerate}
\item  $\bar{\Delta}_1$ is discrete;
\item  $r_1\in \bar{\Delta}_1$;
\item  $\body{r}=\emptyset$ for any $r\in \bar{\Delta}_1$.
\end{enumerate}

Suppose that $\bar{\Delta}_i$ is well-defined and $r_{m_i}$ is the last rule
of $\bar{\Delta}_i$, we define $\bar{\Delta}_{i+1}$
as the largest section of $\langle r_i\rangle_{i\in I}$ satisfying
the following conditions:
\begin{enumerate}
\item  $\bar{\Delta}_{i+1}$ is discrete;
\item  $r_{m_i+1}\in \bar{\Delta}_{i+1}$;
\item  Either $\body{r}\subseteq \{\head{r'}\mid r'\in \cup_{j=0}^i \bar{\Delta}_j\}$ 
or $\head{r}\in \{\head{r'}\mid r'\in \cup_{j=0}^i \bar{\Delta}_j\}$
for any $r\in \bar{\Delta}_{i+1}$.
\item  Disjoint with $\cup_{j=0}^i \bar{\Delta}_j$.
\end{enumerate}

Denote $\bar{X}_i= \{\head{r}\mid r\in \cup_{j=0}^i \bar{\Delta}_j\}$. Then
we observe the following fact: 

{\em If $r\in \bar{\Delta}_{i+1}$ such that $\pbody{r}$ is satisfied by $\bar{X}_{i-1}$
and no rule $r'\in \bar{\Delta}_i$ with $r<r'$, then we can move $r$ from
$\bar{\Delta}_{i+1}$ to $\bar{\Delta}_i$, the resulting sequence of rules
is still $<^\WZL$-preserving.
}

Without loss of generality,  assume that our sequence 
$\langle\bar{\Delta}_i\rangle$ is fully
transformed by the above transformation.
Since $\cup_{i=0}^{\infty} \bar{X}_i=X$, we can prove 
$\cup_{i=0}^{\infty} X_i=X$ by proving $\bar{X}_i=X_i$ for every 
$i\in I$.
Thus, it suffices to prove $\bar{\Delta}_i=\ugr{X_i}$
for every $i\in I$. 
We use induction on $i$:
\paragraph{Base}  $\bar{\Delta}_0=\ugr{X_0} =\emptyset$.
\paragraph{Step}  Assume that 
\(
\bar{\Delta}_i=\ugr{X_i}
\),
we want to prove 
\(
\bar{\Delta}_{i+1}=\ugr{X_{i+1}}
\).
\begin{enumerate}
\item 
\(
\bar{\Delta}_{i+1}\subseteq \ugr{X_{i+1}}
\): 
For any $r_t\in \bar{\Delta}_{i+1}$, by the condition 3 in the 
construction of $\bar{\Delta}_{i+1}$, either $r_t$ 
is active wrt $(X_i,X)$ or $\head{r_t}\in X_i$. 
If $\head{r_t}\in X_i$, it is obvious that $r\in\ugr{X_{i+1}}$.
Thus, we assume that $r_t$ is active wrt $(X_i,X)$. For any $r'$ such that $r_t<r'$
and $r'$ is active wrt $(X,X_i)$ then $\pbody{r'}\subseteq X$
and $r'$ is not defeated by $X_i$. 
By induction, 
$\cup_{k<t}\head{r_k}\subseteq X_i=\bar{X}_i$, thus $r'$ is not defeated by
$\cup_{k<t}\head{r_k}$. 
By Definition~\ref{def:order:preservation:W}, 
it should be the case that $r'\in\GR{\Pi}{X}$, which implies that 
$r'\in\bar{\Delta}_i=\ugr{X_i}$. Therefore, $r'$ is not a \WZL-preventer
of $r_t$ in the context of $(X_i,X)$. That is, $r_t\in \ugr{X_{i+1}}$. 
\item
\(
\ugr{X_{i+1}}\subseteq \bar{\Delta}_{i+1}
\): 
For $r\in \ugr{X_{i+1}}$, we claim that 
\(r\in  \bar{\Delta}_{i+1}
\).
Otherwise, there would exist $t>i+1$ such that 
\(r\in \bar{\Delta}_t
\).
Notice that, by induction assumption, 
\(\pbody{r}\subseteq X_i
\)
(Note that $\head{r}\in X_i$ is impossible because we assume that $r\in\bar{\Delta}_t$
and $t>i+1$).
Thus, it must be the case that there is at least one rule 
\(
r'\in \cup_{j=i+1}^{t-1} \ugr{X_j}
\)
such that $r'<r$. But
$r'$ is active wrt $(X,X_{i+1})$, which contradicts to 
$r\in \ugr{X_{i+1}}$. Therefore, 
\(
\ugr{X_{i+1}}\subseteq \bar{\Delta}_{i+1}
\).
\quad\QED
\end{enumerate} 
\end{interproof}

%
\begin{lemma}\label{lem:results:W:2e3}
  Let $(\Pi,<)$ be an ordered logic program over $\mathcal{L}$
  and let $X$ and $Y$ be consistent sets of literals.   
  Then, we have that
  \begin{enumerate}
  \item if $X$ is a $<^{\WZL}$-preserving answer set of\/ $\Pi$,
    then there is some answer set $Y$ of\/ $\TransW{\Pi,<}$
    such that
    \(
    X=Y\cap\mathcal{L}
    \);
  \item if\/ $Y$ is an answer set of\/ $\TransW{\Pi,<}$,
    then 
   $X$ is a $<^{\WZL}$-preserving.
  \end{enumerate}
\end{lemma}
%
\begin{interproof}{lem:results:W:2e3}
\paragraph{1} 
Let $X$ be a $<^{\WZL}$-preserving answer set of\/ $\Pi$.
Define 
\begin{eqnarray*}
  Y &=&\quad \{\head{r} \mid r\in\GR{\Pi}{X}\}
  \\& &\cup\;\{\applied{n_r}\mid r    \in\GR{\Pi}{X}\}
       \cup  \{\blocked{n_r}\mid r\not\in\GR{\Pi}{X}\}
  \\& &\cup\;\{\ok{n_r}   \mid r\in\Pi\}
       \cup  \{\oko{n_r}{n_{r'}}\mid r,r'\in\Pi\}
  \\& &\cup\;\{\PREC{\name{r}}{\name{r'}}\mid r<r'\}
       \cup  \{\neg (\PREC{\name{r}}{\name{r'}})\mid r\not< r'\}
\end{eqnarray*}

Notice that $L\in X$ iff $L\in Y$.
We want to show that 
\(
Y=\Cn{\redwy}
\) by two steps:

\paragraph{``$\supseteq$''-part}
For any
\(
s\in \mathsf{T}^{\WZL}(\Pi,<)
\), 
if $s^{+}\in \redwy$ and $\pbody{s}\subseteq Y$, 
we need to prove $\head{s}\in Y$ by cases.
\begin{description}
\item[Case 1]\quad 
\(
\ap{1}{r}:\head{r}\LPif\applied{\name{r}}
\).
Since
\(
\ap{1}{r}=\ap{1}{r}^{+}
\), 
\(
\ap{1}{r}\in \redwy
\).
If
\(
\applied{n_r}\in Y
\), 
then
\(
r\in \GR{\Pi}{X}
\).
This implies 
\(
\head{r}\in Y
\).
\item[Case 2]\quad  
\(
\ap{2}{r}:\applied{\name{r}}\LPif\ok{\name{r}},\body{r}
\).
If
\(
\ok{n_r}\in Y
\), 
\(
\pbody{r}\subseteq Y
\)
and
\(
\nbody{r}\cap Y=\emptyset
\),
then
\(
\pbody{r}\subseteq X
\)
and
\(
\nbody{r}\cap X=\emptyset
\). 
This implies that
\(
r\in \GR{\Pi}{X}$ and thus $\applied{n_r}\in Y
\).
\item[Case 3]\quad
\(
\bl{1}{r}{L^{+}}:\blocked{\name{r}}\LPif\ok{\name{r}}, \naf L^{+}
\). 
If
\(
\ok{n_r}\in Y
\)
and
\(
L^{+}\not\in Y
\), 
then
\(
L^{+}\not\in X
\).
That is,
\(
r\not\in \GR{\Pi}{X}
\)
and thus
\(
\blocked{r}\in Y
\).
\item[Case 4]\quad 
\(
\bl{2}{r}{L^{-}}:\blocked{\name{r}}\LPif\ok{\name{r}},L^{-}
\).
If
\(
\ok{n_r}\in Y
\)
and
\(
L^{-}\in Y
\), 
then
\(
L^{-}\in X
\).
That is,
\(
r\not\in \GR{\Pi}{X}
\) 
and thus 
\(
\blocked{r}\in Y
\).
\item[Case 5] \quad For the rest of rules in
\(
\mathsf{T}^{\WZL}(\Pi,<)
\),
we trivially have that
\(
\head{s}\in Y
\)
whenever
\(
s^{+}\in \redwy
\)
and
\(
\pbody{s}\subseteq Y
\).
\end{description}
\paragraph{``$\subseteq$''-part}
Since $X$ is a $<^{\WZL}$-preserving answer set of\/ $\Pi$,
there is an enumeration
\(
\langle r_i \rangle_{i\in I}
\) 
of $\GR{\Pi}{X}$ satisfying all conditions in 
Definition~\ref{def:order:preservation:W}.
This enumeration can be extended to an enumeration of $\Pi$ as follows:

For any
\(
r\not\in \GR{\Pi}{X}
\), 
let $r_i$ be the first rule that blocks
$r$ and $r_j$ be the last rule s.~t. $r<r_j$.
Then we insert $r$ immediately after $r_{max\{i,j\}}$. 
For simplicity, the extended enumeration is still denoted
\(
\langle r_i \rangle_{i\in I}
\).
Obviously, this enumeration has the following property by 
Definition~\ref{def:order:preservation:W}.
\begin{lemma}\label{lem:order:keep:pref}
Let
\(
\langle r_i \rangle_{i\in I}
\)
be the enumeration for $\Pi$ defined as above. 
If
\(
r_i<r_j
\), 
then 
\(
j<i
\).
\end{lemma}
For each $r_i\in \Pi$, we define $Y_i$ as follows:
\[
\begin{array}{clcl}
 & \{\head{r_i},\applied{\name{r_i}}\mid r_i\in\GR{\Pi}{X},i\in I\} &
\cup & \{\blocked{\name{r_i}}\mid r_i\not\in\GR{\Pi}{X},i\in I\} \\
\cup & \{\ok{\name{r_i}}\mid i\in I\}  &                        
  \cup & \{\oko{\name{r_i}}{\name{r_j}}\mid i,j\in I\} \\
\cup & \{\PREC{\name{r}}{\name{r'}}\mid r<r'\} &
\cup & \{\neg(\PREC{\name{r}}{\name{r'}})\mid r\not< r'\}.
\end{array}
\] 

We prove
\(
Y_i\subseteq\Cn{\reduct{\mathcal{T}(\Pi)}{Y}}
\) 
by using induction on $i$.
  \begin{description}
    \item[Base] 
     Consider $r_0\in\Pi$.
     Given that $X$ is consistent,
     we have $r_0\not<r$
     for all $r\in\Pi$
     by Definition~\ref{def:order:preservation:W}(3).
     Thus,
     \(
     \neg (\PREC{\name{r_0}}{\name{r}})\in Y
     \)
     for all $r\in\Pi$.
     Consequently,
     \[
     \reductr{\cok{2}{r_0}{r}}:
     {\oko{\name{r_0}}{\name{r}}}\LPif\
     \in\redwy
     \text{ for all }
     r\in\Pi.
     \]
     This implies
     \(
     {\oko{\name{r_0}}{\name{r}}}\in\Cn{\redwy}
     \)
     for all $r\in\Pi$.
  
     Let $\Pi=\{r_0,r_1,\ldots,r_k\}$.
     Since
     \begin{equation}
    \label{eq:ok:i}
              {\cokt{1}{r_0}}
    = \reductr{\cokt{1}{r_0}}
    : {\ok{\name{r_0}}}
      \LPif
      \begin{array}[t]{l}
        {\oko{\name{r_0}}{\name{r_1}},\dots,\oko{\name{r_0}}{\name{r_k}}}
        \in\redwy,
      \end{array}
  \end{equation}
  thus
  \(
  {\ok{\name{r_0}}}\in\Cn{\redwy}
  \).
  We distinguish two cases.
  \begin{description}
  \item[Case 1]
    If
    \(
    r_0\in\GR{\Pi}{X}
    \), 
    we have 
    \(
    \pbody{r_0}=\emptyset
    \)
    by Definition~\ref{def:order:preservation:W}(1),
    and 
    \(
    \nbody{r_0}\cap X=\emptyset
    \)
    which also implies
    \(
    \nbody{r_0}\cap Y=\emptyset
    \).
    Thus
    \begin{equation}
                \ap{2}{r_0}
      =\reductr{\ap{2}{r_0}}
      :
      \applied{\name{r_0}}\LPif\ok{\name{r_0}}
      \in\redwy
      \ .
    \end{equation}
    Accordingly, we obtain
    \(
    {\applied{\name{r_0}}}\in\Cn{\redwy}
    \)
    by $\ok{\name{r_0}}\in \Cn{\redwy}$.

    Furthermore, from 
    \begin{equation}
                \ap{1}{r_0}
      =\reductr{\ap{1}{r_0}}
      :
      \head{\name{r_0}}\LPif\applied{\name{r_0}}
      \in\redwy
      \ ,
    \end{equation}
   we obtain
    \(
    {\head{r_0}}\in\Cn{\redwy}
    \).
  \item[Case 2]
    If
    \(
    r_0\in\Pi\setminus\GR{\Pi}{X}
    \),
    we must have
    \(
    \pbody{r_0}\not\subseteq X
    \)
    by Definition~\ref{def:order:preservation:W}.
    That is, 
    \(
    \pbody{r_0}\not\subseteq Y
    \).
    %
%
    Then, there is some
    \(
    L^{+}\in\pbody{r_0}
    \)
    with $L^{+}\not\in X$.
    We also have $L^{+}\not\in Y$.
    Therefore,
    \begin{equation}
      \label{eq:bl:base}
                  \bl{1}{r_0}{L^{+}}
        =\reductr{\bl{1}{r_0}{L^{+}}}
        :
        \blocked{\name{r_0}}\LPif\ok{\name{r_0}}
        \in\redwy
        \ .
    \end{equation}
    Since we have shown above that
    $\ok{\name{r_0}}\in\Cn{\redwy}$,
    we obtain
    \[
    \blocked{\name{r_0}}\in\Cn{\redwy}.
    \]      
  \end{description}
    \item[Step]
  Assume that $Y_j\subseteq \redwy$ for all $j<i$, we show
  $Y_i\subseteq \redwy$ by cases.
  \begin{itemize}
    \item 
      \(
      \oko{\name{r_i}}{\name{r_j}}\in\Cn{\redwy} 
      \):
      
    If $r_i<r_j$, then
    \(
    \PREC{\name{r_i}}{\name{r_j}}\in Y
    \)
    and
           $j<i$ by Lemma~\ref{lem:order:keep:pref}. 
           
           By the induction assumption, either 
    \(
    \applied{\name{r_j}}\in\Cn{\redwy}
    \)
    or
    \(
    \blocked{\name{r_j}}\in\Cn{\redwy}
    \).
    Since 
    \(
    \cok{3}{r_i}{r_j}
    \), 
    \(
    \cok{4}{r_i}{r_j}
    \)
    are in $\redwy$,
    we have 
    \[
    \oko{\name{r_i}}{\name{r_j}}\in\Cn{\redwy}
    \qquad
    \text{ whenever }r_i<r_j\ .
    \]
    
    If 
    \(
    r_i\not<r_j
    \), then 
    \(
    \neg (\PREC{\name{r_i}}{\name{r_j}})\in Y
    \)
    and thus
    \[
    \reductr{\cok{2}{r_i}{r_j}}
    :
    {\oko{\name{r_i}}{\name{r_j}}}\LPif\
    \in{\redwy}.
    \]
    Consequently, for all $j\in I$,
    \(
    \oko{\name{r_i}}{\name{r_j}}\in\Cn{\redwy}
    \).
    \item \(
    \ok{\name{r_i}}\in\Cn{\redwy}
    \):
    It is obtained directly by 
    \(
    \reductr{\cokt{1}{r_i}}=\cokt{1}{r_i}\in \Cn{\redwy}.
    \)
    \item If 
    \(
    {r_i}\in\GR{\Pi}{X}
    \), then
    \(
    \{\applied{r_i},\head{r_i}\}\subseteq \Cn{\redwy}.
    \)

    By Definition~\ref{def:order:preservation:W},
    \(
    \pbody{r_i}\subseteq\{\head{r_j}\mid r_j\in\GR{\Pi}{X}, j<i\}
    \)

    or
    \(
    \head{r_i}\in\{\head{r_j}\mid r_j\in\GR{\Pi}{X}, j<i\}.
    \)
    By the induction assumption,
    \(
    \pbody{r_i}\subseteq\Cn{\redwy}.
    \)
    Also, ${r_i}\in\GR{\Pi}{X}$ implies $\nbody{r_i}\cap X=\emptyset$.
    Thus $\nbody{r_i}\cap Y=\emptyset$.
    
    This means that
    \begin{equation}
      \label{eq:ap:ii:ii}
                  \ap{2}{r_i}
        =\reductr{\ap{2}{r_i}}
        :
        \applied{\name{r_i}}\LPif\ok{\name{r_i}},\pbody{r_i}
        \in\redwy.
    \end{equation}
    As shown above, 
    \(
    \ok{\name{r_i}}\in\Cn{\redwy}
    \). 
    Therefore,
    \(
    \applied{\name{r_i}}\in\Cn{\redwy}
    \).
    Accordingly, we obtain
    \(
    {\head{r_i}}\in\Cn{\redwy}
    \)
    due to $\reductr{\ap{1}{r_i}}\in\redwy$.
    
  \item
    If
    \(
    {r_i}\in\Pi\setminus\GR{\Pi}{X}
    \), 
    \(
    \blocked{\name{r_i}}\in \Cn{\redwy}
    \):
    We consider three possibilities.
    \begin{enumerate}
    \item  $\pbody{r_i}\not\subseteq X$:
      then there is some
      \(
      L^{+}\in\pbody{r_i}
      \)
      with $L^{+}\not\in X$.
      
      Also, $L^{+}\not\in Y$.
      Thus,
      \begin{equation}
        \label{eq:bl:ii:ii}
                    \bl{1}{r_i}{L^{+}}
          =\reductr{\bl{1}{r_i}{L^{+}}}
          :
          \blocked{\name{r_i}}\LPif\ok{\name{r_i}}
          \in\redwy
          \ .
      \end{equation}
      By $\ok{\name{r_i}}\in\Cn{\redwy}$,
      we have
      \(
      \blocked{\name{r_i}}\in\Cn{\redwy}
      \).
     \item
      \(
      \nbody{r}\cap\{\head{r_j}\mid r_j\in\GR{\Pi}{X}, j<i\}\neq\emptyset
      \):
      then there is some
      \(
      L^{-}\in\nbody{r_i}
      \)
      with
      \(
      L^{-}\in\{\head{r_j}\mid r_j\in\GR{\Pi}{X}, j<i\}.
      \)
      That is, 
      \(
      L^{-}=\head{r_j}
      \)
      for some $r_j\in\GR{\Pi}{X}$ with $j<i$.
      With the induction hypothesis, we then obtain
      \(
      L^{-}\in\Cn{\redwy}.
      \)
      Since
      \(
      \ok{\name{r_i}}\in\Cn{\redwy},
      \)
      we obtain
      \(
      \blocked{\name{r_i}}\in\Cn{\redwy}.
      \)
     \item 
     $\head{r_i}\in\{\head{r_j}\mid r_j\in\GR{\Pi}{X}, j<i\}
      \): this is obtained directly by the induction assumption.
     \end{enumerate}
   \end{itemize}
  \end{description}
\paragraph{2}
Let $Y$ be a consistent answer set of 
\(
\mathsf{T}^{\WZL}(\Pi,<)
\)
and 
\(
    X=Y\cap\mathcal{L}
    \).
To prove that $X$ is a $<^{\WZL}$-preserving answer set of\/ $\Pi$,
it suffices to prove that the following two propositions $P1$ and $P2$:
\begin{description}
\item[P1] $X$ is an answer set of\/ $\Pi$: that is, $\Cn{\reduct{\Pi}{X}}=X$.
  \begin{enumerate}
  \item $\Cn{\reduct{\Pi}{X}}\subseteq X$:
    Let $r\in \Pi$ s.~t. 
    \(
    \pbody{r}\subseteq X
    \)
    and
    \(
    \nbody{r}\cap X=\emptyset
    \).

    Then 
    \(
    \pbody{r}\subseteq Y
    \) 
    and
    \(
    \nbody{r}\cap Y=\emptyset
    \).
    By Lemma~\ref{lem:compilation:basic}, $\ok{n_r}\in Y$ and thus 
    \(
    \ap{2}{r}^{+}\in \redwy
    \).
    Since $Y$ is closed under $\redwy$,
    \(
    \applied{n_r}\in Y
    \) 
    and thus
    \(
    \head{r}\in Y
    \).
    This means $\head{r}\in X$.

   \item $X\subseteq \Cn{\reduct{\Pi}{X}}$:
     Since 
    \(
    X=Y\cap\mathcal{L}=
    (\cup_{i\geq 0}\Ti{i}{\redwy}{\emptyset}
    \cap\mathcal{L},
    \)
    we need only to show by induction on $i$ that, for $i\ge 0$,
    \begin{equation} \label{eqn:W:1}
      (\Ti{i}{\redwy}{\emptyset}
      \cap\mathcal{L})\subseteq \Cn{\reduct{\Pi}{X}}.
    \end{equation}
   \end{enumerate}
  \begin{description}
    \item[Base] 
    It is obvious that 
    \(
    \Ti{0}{\redwy}{\emptyset}=\emptyset.
    \) 
    \item[Step]
    Assume that (\ref{eqn:W:1}) holds for $i$, we want to prove
  (\ref{eqn:W:1}) holds for $i+1$. 
  
  If $L\in \Ti{i+1}{\redwy}{\emptyset}$,
  then there is a rule $r\in \Pi$ s.~t. $\head{r}=L$,
  \(
  \ap{1}{r}^{+}, \ap{2}{r}^{+} \in \redwy
  \)
  and 
   \(
   \{\applied{n_r},\ok{n_r}\}\cup \pbody{r}\subseteq 
           \Ti{i}{\redwy}{\emptyset}.
   \)
This also means $\nbody{r}\cap Y=\emptyset$.
By the induction assumption, $\pbody{r}\in \Cn{\reduct{\Pi}{X}}$.
Together with $\nbody{r}\cap X=\emptyset$, we have
$r\in \reduct{\Pi}{X}$ and thus $\head{r}\in \Cn{\reduct{\Pi}{X}}$.
Therefore, $X=\Cn{\reduct{\Pi}{X}}$.
  \end{description}
\item[P2] $X$ is $<^{\WZL}$-preserving:
Since $Y$ is a standard answer set of $\mathsf{T}^{\WZL}(\Pi,<)$,
there is a grounded enumeration
\(
\langle s_k\rangle_{k\in K}
\)Induction
of $\GR{\mathsf{T}^{\WZL}(\Pi)}{Y}$.
Define
\(
\langle r_i\rangle_{i\in I}
\)
as the enumeration obtained from
\(
\langle s_k\rangle_{k\in K}
\)
by
\begin{itemize}
\item deleting all rules apart from those of form
  \ap{2}{r},
  \bl{1}{r}{L^{+}},
  \bl{2}{r}{L^{-}};
\item replacing each rule of form
  \ap{2}{r},
  \bl{1}{r}{L^{+}},
  \bl{2}{r}{L^{-}}
  by $r$;
\item removing duplicates\footnote{Duplicates can only occur if a rule is blocked
    in multiple ways.} by increasing $i$.
\end{itemize}
for $r\in\Pi$ and $L^{+}\in\pbody{r}$, $L^{-}\in\nbody{r}$.

We justify that the sequence 
\(
\langle r_i\rangle_{i\in I}
\)
satisfies the conditions in Definition~\ref{def:order:preservation:W}:
\begin{enumerate}
\item Since
\(
\langle s_k\rangle_{k\in K}
\)
is grounded, Condition 1 is satisfied.
\item If $r_i<r_j$, we want to show $j<i$.
Since $\oko{n_i}{n_j}\in Y$, at least one of   
  \ap{2}{r_j},
  \bl{1}{r_j}{L^{+}},
  \bl{2}{r_j}{L^{-}}
appears before any of
  \ap{2}{r_i},
  \bl{1}{r_i}{L^{+}},
  \bl{2}{r_i}{L^{-}}.
Thus, $j<i$.
\item Let $r_i<r'$ and $r'\in \Pi\setminus \GR{\Pi}{X}$.
Suppose that $\pbody{r}\subseteq X$ and 
$\head{r}\not\in \{\head{r_k} \mid k<i\}$. 
Since $\nbody{r}\cap X\neq \emptyset$, there is some $L^{-}\in\nbody{r}$
s.~t. $L^{-}\in X$. Then $L^{-}\in Y$. Without loss of generality,
let $L^{-}$ is included in $Y$ through rule $s_{k_0}$.
Furthermore, we can assume that there is no $k'<k_0$ such that
$s_{k'}$ is before $s_{k_0}$, $\head{s_{k'}}\in \nbody{r}$ and
$\head{s_{k'}}\in X$.
Since $\ok{r_i}\in Y$, we have $\oko{n_i}{n_{r'}}$. 
This implies,
$\blocked{n_{r'}}\in Y$ and $\bl{2}{r_i}{L^{-}}$ 
appears before $\ap{2}{r}$ in 
\(
\langle s_k\rangle_{k\in K}
\).
Thus, $L^{-}\in \{\head{r_k}\mid k<i\}$.
\quad\QED
\end{enumerate}
\end{description}
\end{interproof}


%

%
\begin{shortproof}{thm:CP:B}
See the proof of Theorem~\ref{thm:results:B}.
\quad\QED
\end{shortproof}

\begin{shortproof}{thm:stratified:perfect:B}
Similar to Proof~\ref{thm:stratified:perfect:W:i}.
\quad\QED
\end{shortproof}

\begin{myproof}{thm:results:B}
Throughout the proofs for Theorem~\ref{thm:results:B},  
the set $X_i$ for any $i\ge 0$ is defined 
as in Definition~\ref{def:be:operator}.
By the definition of $\mathcal{E}_X(\Pi,<)$, we observe the following
facts:
\begin{description}
\item[F1] $X$ is a standard answer set of $(\Pi,<)$ iff 
$X$ is a standard answer set of $\mathcal{E}_X(\Pi,<)$.
\item[F2] $X$ is a $<^{\BE}$-preserving answer set of $(\Pi,<)$ iff 
$X$ is a $<^{\BE}$-preserving answer set of $\mathcal{E}_X(\Pi,<)$.
\item[F3] $X$ is a standard answer set of $\TransB{\Pi,<}$ iff 
$X$ is a standard answer set of $\Trans{\mathcal{E}_X(\Pi,<)}$.
\end{description}
Having the above facts, we can assume that
\(
(\Pi,<)=\mathcal{E}_X(\Pi,<)
\).
Thus,
we need only to prove the following Lemma~\ref{lem:result:iiia}
and Lemma~\ref{lem:result:iiib}.
\quad\QED
\end{myproof}
Given a statically ordered logic program $(\Pi,<)$ and
a set $X$ of literals, set
\renewcommand{\TPo}[3]{\ensuremath{(\mathcal{T}^{\BE})_{(#1,#2),#3}}}
\renewcommand{\TPDiO}[4]{\TPo{#2}{#3}{#4}^{#1}}
%
\(
X_i=\TPDi{i}{\Pi}{<}{X}{\emptyset}
\)
for $i\ge 0$.
\begin{lemma}\label{lem:result:iiia}
  Let $(\Pi,<)$ be a statically ordered logic program over $\mathcal{L}$
  and let $X$ be an answer set of\/ $\Pi$.
  %
  Then, the following propositions are equivalent.
  \begin{enumerate}
   \item $X$ is a \BE-preferred answer set of\/ $(\Pi,<)$;
  \item
    \(
    \mathcal{C}''_{(\Pi_X,<_X)}(X)= X
    \).
  \end{enumerate}
\end{lemma}
%
To prove Lemma~\ref{lem:result:iiia}, some preparations are in order.
%
\begin{definition}\label{def:afp:breiter'}
Let $(\Pi,<)=\langle r_1,r_2,\ldots,r_n\rangle$ be a totally ordered 
logic program, where $r_{i+1}<r_i$ for each $i$, and let $X$ be a 
set of literals.

We define
\begin{eqnarray*}
  {\bar X}_0    &=& \emptyset\qquad\text{ and for }i \geq 0
  \\
  {\bar X}_{i+1}&=&{\bar X}_i
            \cup 
            \left\{\head{r_{i+1}}\left|\;
\if{\right.\right.\\&&\left.\left.}\fi
                           \begin{array}{ll}
                            (1) & r_{i+1} \text{ is active wrt } (X,X)
                                  \text{ and}
                            \\
                            (2) & \text{there is no rule }r'\in\Pi
                                  \text{ with } r_{i+1}< r'
                            \\  & \text{such that}
                            \\
                                & (a)\ r'
                                      \text{ is active wrt } (X,{\bar X}_i)
                                      \text{ and}
                            \\
                                & (b)\ \head{r'}\not\in {\bar X}
                          \end{array}
                          \right\}\right.
\end{eqnarray*}
Then,
\(
{\mathcal D}_{(\Pi,<)}(X)=\bigcup_{i\geq 0} {\bar X}_i
\)
if $\bigcup_{i\geq 0} {\bar X}_i$ is consistent.
Otherwise, ${\mathcal D}_{(\Pi,<)}(X)=Lit$.
\end{definition}
If we want to stress that ${\bar X}_i$ is for ordering $<$,
we will also write it as ${\bar X}^{<}_i$. We assume the same
notation for $X_i$.
\begin{lemma} \label{lemma:def:BE}
Let $(\Pi,<)$ be an ordered logic program. $\bar{X}_i$ for $i\ge 0$ is 
given as above and $\Pi$ is prerequisite-free.
Then $X_i=\bar{X}_{k_i}$ for some non-decreasing sequence $\{k_i\}_{i\ge 0}$
with
\(
0\le k_1\le\cdots\le k_i\le\cdots
\).
\end{lemma}
\begin{interproof}{lemma:def:BE}
Without loss of generality, assume that 
\(
{\bar X}_0=\cdots ={\bar X}_{k_1},
\)
\(
{\bar X}_{k_1+1}=\cdots ={\bar X}_{k_2},
\)
\(
\ldots.
\)
Then by a simple induction on $i$, we can directly prove that
\[
X_0={\bar X}_0,
X_1={\bar X}_{k_1+1},
\ldots,
X_i={\bar X}_{k_i+1},
\ldots.
\quad\QED
\]
\end{interproof}
\begin{lemma} \label{BE:total+prerequisite-free}
The conclusion of Lemma~\ref{lem:result:iiia} is correct for
ordered logic program $(\Pi,<)$ if $\Pi$ is prerequisite-free
and $<$ is total.
\end{lemma}
\begin{interproof}{BE:total+prerequisite-free}
Since $\Pi$ is prerequisite-free, we have that $\Pi_X=\Pi$.
By Lemma \ref{lemma:def:BE}, it is enough to prove that
$X=\cup {\bar X}_i$ iff $X=\cup X_i$ (see Definition~\ref{def:be:operator}).
For simplicity, we say a rule
$r$ is applicable wrt $(X,{\bar X}_i)$ (only in this proof) if $r$ satisfies
the conditions in the definition of ${\bar X}_{i+1}$.
\paragraph{if part} If $X=\cup {\bar X}_i$, we want to prove that
$X=\cup X_i$. It suffices to show that
${\bar X}_i=X_i$ hold for all $i\ge 0$. We use induction on
$i\ge 0$:
   \begin{description}
     \item[Base] ${\bar X}_0=X_0=\emptyset$.
     \item[Step] Assume that ${\bar X}_{i-1}=X_{i-1}$, we need to
show that ${\bar X}_i=X_i$.
      \begin{enumerate} 
      \item
        ${\bar X}_i\subseteq X_i$: 

        If ${\bar X}_i={\bar X}_{i-1}$, the inclusion follows from the induction assumption;

        If ${\bar X}_i\neq {\bar X}_{i-1}$, then $r_i$ is applicable wrt $(X,{\bar X}_{i-1})$.

        Thus, $r_i$ is not defeated by $X$ by Definition~\ref{def:afp:breiter'}. 

%
      \item 
        $X_i\subseteq {\bar X}_i$:
        If $X_i=X_{i-1}$, the inclusion follows from the induction assumption;
        Let $X_i\neq X_{i-1}$, that is, $head(r_i)\in X_i$.
        Then we can assert that $head(r_i)\in {\bar X}_i$.

        Otherwise, if $head(r_i)\not\in {\bar X}_i$, there will be two
        possible cases because $\Pi$ is prerequisite-free:
         \begin{itemize}
         \item
           $r_i$ is not active wrt $(X,X)$:
           then there exists a literal $l\in body^{-}(r_i)$ such that
           $l\in X$. 
           On the other hand, since $\head{r_i}\in X_i$, $r_i$ is not defeated by
           $X_{i-1}={\bar X}_{i-1}$, so we have $l\not\in {\bar X}_{i-1}$. 
           This implies that there exists $t\le i$ such that 
           \(
           l\in {\bar X}_t\setminus {\bar X}_{i-1}.
           \)
           Thus, $l=head(r_t)$ and $r_t < r_i$.
           Notice that $r_i$ is active wrt $(X,X_{i-1})=(X,{\bar X}_{i-1})$
           and $head(r_i)\not\in X$, thus 
           $r_i$ is active wrt $(X,{\bar X}_{t-1})$ and 
           $head(r_i)\not\in {\bar X}_{t-1}$.
           This implies that $r_i$ is a preventer of $r_t$. 
           Therefore, 
           $head(r_t)\not\in {\bar X}_t$ and so by $X=\cup\bar{X}_i$,
           $head(r_t)\not\in {\bar X}$, contradiction.
         \item There is a rule $r'\in\Pi$ with $r_i<r'$ such that
          $r'$ is active wrt $(X,\bar{X}_{i-1})$ and $\head{r'}\not\in X$.
          Since there are only a finite number of rules in $\Pi$ which are preferred
          over $r_i$, so this case is impossible.
         \end{itemize}
     Combining the  two cases, we have
     $X_{i+1}\subseteq {\bar X}_{i+1}$.
     Thus, $X_i= {\bar X}_i$ for all $i\ge 0$. 
   \end{enumerate} 
\end{description}
\paragraph{only-if part} 
Suppose that $X=\cup X_i$ and $X$ is an answer set of $\Pi$,
we want to prove that 
\[
X=\cup {\bar X}_i:
\]
\begin{enumerate}
\item 
We prove ${\bar X}_i\subseteq X$ by using induction on $i$.
   \begin{description}
   \item[Base] ${\bar X}_0=\emptyset\subseteq X$. 
   \item[Step] Assume that ${\bar X}_i\subseteq X$.
     If $head(r_{i+1})\in {\bar X}_{i+1}$, then $r_{i+1}$ is not
     defeated by $X$ and thus not defeated by $X_i$.
     Thus $head(r_{i+1})\in X_{i+1}$.                    
   \end{description}
\item
$X\subseteq \cup {\bar X}_i$: it is sufficient to show
that $X_i\subseteq {\bar X}_i$ by using induction on $i$.
  \begin{description}
   \item[Base] $X_0=\emptyset={\bar X}_0$.
   \item[Step] Assume that $X_k\subseteq {\bar X}_k$ for $k\le i$,
     then we claim that ${\bar X}_i=X_i$.
     On the contrary, assume that
     $head(r_{i+1})\in X_{i+1}\setminus {\bar X}_{i+1}$.
     From $X=\cup X_i$, we have $head(r_{i+1})\in X$.
     Notice that $X$ is an answer set of $\Pi$, so we can
     further assume that $r_{i+1}$ is active wrt $(X,X)$.  
     Therefore, 
     \(
     head(r_{i+1})\not\in {\bar X}_{i+1}
     \) 
     implies that 
     there is a number $t\le i$ such that
     $r_t$ is active wrt $(X,{\bar X}_i)$ but $head(r_t)\not\in {\bar X}_i$.
     Thus, $r_t$ is active wrt $(X,X_{t-1})$ by induction.
     This forces $head(r_t)\in X$ and $r_t$ is not active wrt
     $(X,X)$, contradiction.
     \quad\QED
\end{description}
\end{enumerate}
\end{interproof}
%
\begin{lemma} \label{BE:partial+prerequisite-free}
The conclusion of Lemma~\ref{lem:result:iiia} is correct for
ordered logic program $(\Pi,<)$ if $\Pi$ is prerequisite-free
and $<$ is a partial ordering.
\end{lemma}
\begin{interproof}{BE:partial+prerequisite-free}
\paragraph{if part} 
 Suppose that $X$ is an answer set of $\Pi$ and
$X=\cup {\bar X}_i^{<}$. 

Let $<_t$ be any total ordering on $\Pi$ satisfying
the following three conditions:
\begin{enumerate}
\item 
If $r<r'$ then $r<_tr'$; and
\item
If  $r$ and $r'$ are unrelated wrt $<$ two rules  and they
are applied in producing
${\bar X}_i$ and ${\bar X}_j$ respectively ($i<j$), then $r'<_tr$.
\item
If 
\begin{itemize}
\item $r$ is active wrt $(X,X)$ and
\item $r'$ is active wrt $(X,{\bar X}_i)$ with $head(r')\not\in {\bar X}_i$ for some $i$
and
\item $r$ and $r'$ are unrelated wrt $<$,
\end{itemize} 
then $r'<_t r$.
\end{enumerate}
Notice that the above total ordering $<_t$ exists. 
We want to prove that $X=\cup {\bar X}_i^{<_t}$. By the condition (3) above,
there will be no new preventer in $(\Pi,<_t)$ for any rule $r$
though there may be more rules that are preferred than $r$.
Thus, $\cup {\bar X}_i^{<_t}=\cup {\bar X}_i^{<}$. That is, $X=\cup {\bar X}_i^{<_t}$.
Since $<_t$ is a total ordering, $X=\cup X_i^{<_t}$.
Therefore, $X$ is a BE-preferred answer set of\/$(\Pi,<)$.
\paragraph{only-if part} 
Suppose that $X$ is a BE-preferred answer set
of 
\/$(\Pi,<)$, then there is a total ordering $<_t$ such that
$X=\cup X_i^{<_t}$. 
By Lemma \ref{def:order:preservation:B},
$X=\cup {\bar X}_i^{<_t}$. 
We want to prove that
$\cup {\bar X}_i^{<_t}=\cup {\bar X}_i^{<}$: On the contrary, assume that this
is not true. 
Then $\cup {\bar X}_i^{<_t}\subset \cup {\bar X}_i^{<}$.
That is, there is a rule $r\in \Pi$ such that $head(r)\not\in\cup {\bar X}_i^{<_t}=X$
but $r$ is active wrt $(X,X)$. 
On the other hand, since $X$ is an answer set of $\Pi$,
$head(r)\in X$, contradiction. 
Therefore, $X=\cup {\bar X}_i^{<}$.
\quad\QED
\end{interproof}
\begin{interproof}{lem:result:iiia}
If $\Pi$ is transformed into $\mathcal{E}_X(\Pi)$, then $\Pi$ may be
performed two kinds of transformations:
\begin{enumerate}
\item Deleting every rule having prerequisite $l$ such that
$l\in X$: this kind of rule can be neither active
wrt $(X,X)$ nor a preventer of another rule because it is
not active wrt $(X,{\bar X}_i)$ for any $i\ge 0$.
\item Removing from each remaining rule $r$  all prerequisites.
\end{enumerate}
Suppose that $r$ is changed into $r'$ by this transformation.
Then
\begin{itemize}
\item
$r$ is active wrt $(X,X)$ iff $r'$ is active wrt $(X,X)$;
\item
$r$ is a preventer in $(\Pi,<)$ iff $r'$ is a preventer in
$\mathcal{E}_X(\Pi),<)$. 
\end{itemize}
By Lemma \ref{BE:partial+prerequisite-free}, Lemma~\ref{lem:result:iiia} is proven.
\quad\QED
\end{interproof}
%
\begin{lemma}\label{lem:result:iiib}
  Let $(\Pi,<)$ be a statically ordered logic program over $\mathcal{L}$
  and let $X$ be an answer set of $\Pi$.
  Then $X$ satisfies the Brewka/Eiter criterion for $\Pi$
    (or equivalently for $\mathcal{E}_X(\Pi)$)
    according to~\cite{breeit99a}
if and only if  
    $X$ is a $<_{\BE}$-preserving answer set of $\Pi$.
\end{lemma}
%
To prove this theorem, the following result given in \cite{breeit99a}
is required.
\begin{lemma} \label{lemma:BE}
Let $(\Pi,<)$ be a statically ordered logic program over $\mathcal{L}$
  and let $X$ be an answer set of $\Pi$. 
Then $X$ is a \BE-preferred answer set
if and only if, for
each rule $r\in\Pi$ with $body^{+}(r)\subseteq X$ and
$head(r)\not\in X$, there is a rule $r'\in \GR{\Pi}{X}$ such that
$r<r'$ and $head(r')\in body^{-}(r)$. 
\end{lemma}
\begin{interproof}{lem:result:iiib}
\paragraph{if part}
Let $X$ be a $<_{\BE}$-preserving answer set of $\Pi$.

Assume that $X$ is not a \BE-preferred answer set, by Lemma~\ref{lemma:BE},
then there is a rule $r\in \Pi$ such that the followings hold:
   \begin{enumerate}
   \item 
     $body^{+}(r)\subseteq X$;
   \item
     $head(r)\not\in X$ and
   \item
     For any rule $r'\in \GR{\Pi}{X}$ with $r<r'$,
     $head(r')$ does not defeat $r$. 
   \end{enumerate}
Then,
$head(r')\not\in body^{-}(r)$.
Thus $r'\in \Pi\setminus \GR{\Pi}{X}$.
This contradict to the Condition 2 in 
Definition~\ref{def:order:preservation:B}.
Therefore, $X$ is a \BE-preferred answer set of $\Pi$. 
\paragraph{only-if part}
Suppose that $X$ is a \BE-preferred answer set of $\Pi$. 
Then
$X$ is also a \BE-preferred answer set of $(\Pi,<')$
where $<'$ is a total ordering and compatible with $<$.
Notice that the ordering $<'$ actually determines an
enumeration $\langle r_i \rangle_{i\in I}$ of $\GR{\Pi}{X}$
such that $r_i<'r_j$ if $j<i$.
Thus, this enumeration of $\GR{\Pi}{X}$ obviously satisfies
the condition 1 in Definition~\ref{def:order:preservation:B}.

We prove the Condition 2 is also be satisfied.
Let $r_i<r'$ and $r'\in \Pi\setminus \GR{\Pi}{X}$.
Suppose that $body^{+}(r')\subseteq X$ and $head(r')\not\in X$.
By Lemma~\ref{lemma:BE}, there is a rule $r_j\in \GR{\Pi}{X}$
such that $r'<r_j$ and $head(r_j)\in body^{-}(r')$. 
Thus, the Condition 2 is satisfied.
\quad\QED
\end{interproof}

%
\begin{myproof}{thm:D:W:eq}
Under the assumption of the theorem, we can see that
\(
\TPDX{\Pi}{<}{Y}{X}=\TPWX{\Pi}{<}{Y}{X}
\) 
for any sets $X$ and $Y$ of literals, which implies 
\(
\CPDX{\Pi}{<}{X}=\CPWX{\Pi}{<}{X}
\)
for any set $X$ of literals.
Thus, the conclusion is obtained by Theorem~\ref{thm:results:D}.
\quad\QED
\end{myproof}

\begin{myproof}{thm:D:W}
By comparing Condition II(b) in Definition~\ref{def:Tp:W} and~\ref{def:Tp:D},
we get 
\[
\TPD{\Pi}{<}{Y}{X}\subseteq\TP{\Pi}{<}{Y}{X}.
\]
This means
\(
\CPDXdefault\subseteq\CPXdefault.
\)
If $X$ is a \DST-preferred answer set of $(\Pi,<)$,
it follows from Theorem~\ref{thm:results:D} that
\(
\CPDXdefault=X
\). 
Thus, $X\subseteq\CPXdefault$. 
On the other hand, since a \DST-preferred answer set is also a standard answer set,
we have
\(
\CPXdefault\subseteq C_{\Pi}X=X.
\)
Therefore, $X=\CPXdefault.$
\quad\QED
\end{myproof}

\begin{myproof}{thm:W:B}
By comparing Condition I in Definition~\ref{def:Tp:W} and~\ref{def:Tp:B},
we get 
\[
\TP{\Pi}{<}{Y}{X}\subseteq\TPB{\Pi}{<}{Y}{X}.
\]
This means
\(
\CPXdefault\subseteq\CPBXdefault.
\)
If $X$ is a \WZL-preferred answer set of $(\Pi,<)$,
then
\(
X=\CPXdefault
\). 
Thus, $X\subseteq \CPBXdefault$.
On the other hand, since $X$ is also a standard answer set,
we have
\(
\CPBXdefault\subseteq C_{\Pi}X=X.
\) 
Therefore, $X=\CPXdefault$.
\quad\QED
\end{myproof}

\begin{shortproof}{thm:D:W:B}
It follows directly from Theorem~\ref{thm:D:W} and~\ref{thm:W:B}.
\quad\QED
\end{shortproof}

\begin{myproof}{thm:D:W:B:stratified}
By Theorem~\ref{thm:stratified:perfect:D}, the ordered program
$(\Pi,<_s)$ has the unique \DST-preferred answer set $X^\star$.
Since $<\;\subseteq\; <_s$, $X^\star$ is also a \DST-preferred answer
set
of $(\Pi,<)$. On the other hand, each stratified logic program has
the unique answer set (the perfect model), ie.
$\mathcal{AS}(\Pi)=\{X^\star\}$.
By Theorem~\ref{thm:D:W:B}, we arrive at the conclusion of
Theorem~\ref{thm:D:W:B:stratified}.
\quad\QED
\end{myproof}


\bibliographystyle{acmtrans}

\end{document}